\documentclass[11pt,letterpaper]{article}
\usepackage{amsfonts}
\usepackage{amsmath}
\usepackage{amsthm}
\usepackage{amssymb}
\usepackage{geometry}
\usepackage{indentfirst}
\usepackage{changepage}
\usepackage{graphicx}
\usepackage{epstopdf} 
\usepackage{color}
\usepackage{url}            
\usepackage{mathtools}
\usepackage{algorithm}
\usepackage{algorithmic}
\usepackage{subfigure}
\usepackage{array}
\usepackage{floatflt}
\usepackage{multirow}
\usepackage{wrapfig}
\usepackage{microtype}  
\title{\textbf{Safe Screening for Sparse Conditional Random Fields}}
\author{Weizhong Zhang$^{1}$, Shuang Qiu$^{2}$\thanks{Corresponding author}\\
	$^1$The Hong Kong University of Science and Technology\\
	$^2$University of Michigan, Ann Arbor
}
\date{}

\def \a {\mathbf{a}}
\def \b {\mathbf{b}}

\def \p {\mathbf{p}}

\def \u {\mathbf{u}} 

\def \w {\mathbf{w}}
\def \x {\mathbf{x}}
\def \y {\mathbf{y}}
\def \z {\mathbf{z}}

\def \X {\mathbf{X}}

\def \R {\mathbb{R}}
\def \calA {\mathcal{A}}
\def \calB {\mathcal{B}}
\def \calC {\mathcal{C}}

\def \calF {\mathcal{F}}
\def \calG {\mathcal{G}}
\def \calH {\mathcal{H}}

\def \calJ {\mathcal{J}}

\def \calS {\mathcal{S}}

\def \calY {\mathcal{Y}}

\def \hcalF {\hat{\mathcal{F}}}

\newtheorem{theorem}{Theorem}
\newtheorem{lemma}{Lemma}

\newtheorem{remark}{Remark}

\renewcommand{\eqref}[1]{Eq.~(\ref{#1})}

\begin{document}
	\maketitle	
\begin{abstract} 
	Sparse Conditional Random Field (CRF) is a powerful technique in  computer vision and natural language processing for structured prediction. However, solving sparse CRFs in large-scale applications remains challenging. In this paper, we propose a novel safe dynamic screening method that exploits an accurate dual optimum estimation to identify and remove the irrelevant features during the training process. Thus, the problem size can be reduced continuously, leading to great savings in the computational cost without sacrificing any accuracy on the finally learned model. To the best of our knowledge, this is the first screening method which introduces the dual optimum estimation technique---by carefully exploring and exploiting the strong convexity and the complex structure of the dual problem--- in static screening methods to dynamic screening. In this way, we can absorb the advantages of both the static and dynamic screening methods and avoid their drawbacks. Our estimation would be much more accurate than those developed based on the duality gap, which contributes to a much stronger screening rule. Moreover, our method is also the first screening method in sparse CRFs and even structure prediction models.  Experimental results on both synthetic and real-world datasets demonstrate that the speedup gained by our method is significant.
\end{abstract} 

\section{Introduction}\label{sec:introduction}
Sparse Conditional Random Field (CRF) \cite{lafferty2001conditional, qian2009sparse,sokolovska2010efficient} is a popular technique that can simultaneously perform the structured prediction and feature selection via $\ell_1$-norm penalty. Successful applications have included bioinformatics \cite{liu2006protein, sato2005rna}, text mining \cite{settles2005abner, sha2003shallow} and image processing \cite{he2004multiscale, kumar2004discriminative}. A lot of algorithms \cite{sokolovska2010efficient, pal2006sparse, schmidt2015non} have been developed to efficiently train a sparse CRF model. However, applying sparse CRF to large-scale real problems with a huge number of samples and extremely high-dimensional features remains challenging. 

Screening \cite{EVR:10} is an emerging technique, which has been shown powerful in accelerating the training process of large-scale sparse learning models. The motivation is that all the sparse learning models have a common feature, i.e., their optimal primal or dual solutions always have massive zero valued coefficients, which implies that the corresponding features or samples are irrelevant to the final learned model.  Screening methods accelerate the training process by identifying and removing the irrelevant features or samples during or before the training process. Therefore, the problem size can be reduced dramatically, which leads to substantial savings in the computational cost and memory usage. In recent few years, many specific screening methods for most of the traditional sparse models have been developed, such as Lasso \cite{tibshirani2012strong, wang2013lasso}, tree guided group Lasso \cite{wang2015multi}, sparse logistic regression \cite{wang2014safe}, SVM \cite{ogawa2014safe, shibagaki2016simultaneous,zhang2016scaling} and $\ell_1$-regularized Ising \cite{NIPS20176674}. According to when the screening rules are triggered, they can be classified into two categories, i.e., static screening if triggered before training and dynamic screening  if during training. A nice feature is that most of them are safe in a sense that they achieve the speedup without sacrificing any accuracy. Moreover, since screening rules are developed independently to the training algorithms, they can be integrated with most algorithms flexibly. Experimental studies indicate that the speedups they achieved can be orders of magnitude.

However,  we notice that both of the static and dynamic screening methods have their own advantages and disadvantages. To be precise, static screening rules have a much lower time cost than dynamic screening rules since they only need to be triggered once before training, but they cannot exploit the new information during the training process, such as the current solution would get closer to the optimum and the duality gap would become smaller. Dynamic screening methods can make use of the new information since they are triggered for many times during the training process. However, their performances are always inferior to static screening methods, which is reported in several existing works, such as \cite{zhang2016scaling}. The reason is that dynamic screening always uses a dual optimum estimation developed based on duality gap, which can always be large in the early stage, while in static screening methods, the estimations are always developed using the strong convexity and the variational inequality of the objective functions. This makes the estimation in dynamic screening inaccurate and finally leads to ineffective screening rules. In addition, to the best of our knowledge, specific screening rules for structured prediction models, such as sparse CRF \cite{lafferty2001conditional,sokolovska2010efficient} and structured Support Vector Machine \cite{tsochantaridis2004support},  are still notably absent.

In this paper, we take sparse CRF as an example of structured prediction models and try to accelerate its training process by developing a novel safe dynamic screening rule, due to its popularity in various real applications and the strong craving for scaling up the training in huge problems. We aim to develop a new effective screening rule for sparse CRF, which can absorb the advantages and avoid the disadvantages of the existing static and dynamic screening methods. Our major technical contribution is that we extend the techniques of dual optimum estimation used in static screening methods to dynamic screening. To be specific, we first develop an accurate dual estimation in Section \ref{sec:dual-estimation} by carefully studying the strongly convexity and the dual problem structure, which is only used in the existing static screening methods. Compared with the estimation in static screening methods, our estimation can exploit the new information brought by the current solution. Thus, it would be more and more accurate as the training process goes on.  Different from  dynamic screening methods, our approach does not depend on the duality gap and could be more accurate than that based on duality gap. Then, in Section \ref{sec:screening-rule}, we derive the detailed screening rule based on the estimation by solving a convex optimization problem, which we show has a closed-form solution. At last, we present a framework in Section \ref{sec:screening-rule} to show that our screening rule can be integrated with proper training algorithms in a dynamic manner flexibly. Experimental results (Section \ref{sec:experiments}) on both synthetic and real datasets demonstrate the significant speedup gained by our method. For the convenience of presentation, we postpone detailed proofs of theoretical results in the main text to the supplementary material.

\textbf{Notations:} $\|\cdot\|_1$ and $\|\cdot\|_2$ are the $\ell_1$ and $\ell_2$ norms of a vector, respectively. Let $\langle \x, \y \rangle$ be the inner product of two vectors $\x$ and $\y$. We denote the value of the $j$-th coordinate of the vector $\x$ as $[\x]_j$. Let $[d]=\{1,...,d\} $ for a positive integer $d$. $A\cup B$ is the union of two sets $A$ and $B$. $A \subseteq B$ means that $A$ is a subset of $B$, potentially being equal to $B$. For a subset $\calJ = \{j_1,...,j_k\}$ of $[d]$, we denote $|\calJ|$ to be its cardinality. Moreover, given a matrix $\X$, we denote ${}_\calJ[\X]= (\X_{j_1,:}, ...,\X_{j_k,:})$, where $\X_{j,:}$ is the $j$-th row of $\X$. For a vector $\x$, we denote $[\x]_{\calJ}=([\x]_{j_1},...,[\x]_{j_k})^T$. At last, we denote $\R_+^n$ as the set $\{ \x \in \R^n: [\x]_j >0, j = 1,...,n\}$.  

\section{Basics and Motivations}
In this section, we briefly review some basics of sparse CRF\cite{priol2017adaptive,sokolovska2010efficient} and then motivate our screening rule via KKT conditions. Specifically, we focus on the CRF with an elastic-net penalty, which takes the form of 
\begin{align}
\min_{\w \in \R^d} P(\w;\alpha, \beta) = \frac{1}{n}\sum_{i=1}^{n}\phi_i(-A_i^T \w) + \beta r_{\alpha}(\w), \tag{P} \label{eqn:primal-problem}
\end{align}
where the penalty $r_{\alpha}(\w) = \frac{\alpha}{2}\|\w\|_2^2 + \|\w\|_1$ with $\alpha, \beta$ are two positive parameters, $A_i$ is a $d\times |\calY_i|$ matrix whose columns are the corrected features $\big\{\psi_i(\y) := F(\x_i, \y_i)-F(\x_i, \y)\big \}_{\y\in \calY_i}$ of the feature vector $F(\x,\y) \in \R^d$ defined on the input $\x$ and the structured output $\y \in \calY_i$, the loss $\phi_i(\z)$ is the log-partition function, i.e., $\phi_i(\z) = \log\big(\sum_{j=1}^{|\calY_i|} \exp\big([\z]_j)\big)$ for $\z\in \R_+^{|\calY_i|}$, and $\w$ is the parameter vector to be estimated. We first present the Lagrangian dual problem of (\ref{eqn:primal-problem}) and the corresponding KKT conditions in the theorem below, which plays a fundamentally important role in developing our screening rule. 
\begin{theorem}\label{theorem:primal-dual-kkt}
	Let $\Delta_{k}=\{\u \in \R^k: 0<[\u]_i<1, \sum_{i=1}^k [\u]_i<1 \}$ and $\calS_{\beta}(\cdot)$ be the soft-thresholding operator \cite{hastie2015statistical}, i.e., $[\calS_\beta(\u)]_{i} = \mbox{\rm sign}([\u]_i)(|[\u]_i|-\beta)_{+}$. For any matrix $A\in \R^{p\times q} $ and vector $\b \in \R^{q-1}$, we define $A\circ \b = [A]_{:,1:q-1}\b + (1-\sum_{i=1}^{q-1}[\b]_i) [A]_{:,q}$. Then, the followings hold:\\
	\textup{(i): }The dual problem of (\ref{eqn:primal-problem}) is
	\begin{align}
	&\min_{\theta} D(\theta;\alpha, \beta) = \frac{1}{2\alpha\beta}\|\calS_{\beta}\big(\frac{1}{n}\sum_{i=1}^{n} A_i\circ \theta_i\big)\|_2^2-\frac{1}{n}\sum_{i=1}^{n}H(\theta_i), \tag{D} \label{eqn:dual-problem}\\
	&\mbox{s.t. } \theta = (\theta_1,...,\theta_n) \in \Delta_{|\calY_1|-1} \times...\times \Delta_{|\calY_n|-1}, \nonumber 
	\end{align}
	where $H(\theta_i) = -\sum_{j=1}^{|\calY_i|-1}[\theta_i]_j\log([\theta_i]_j)-(1-\langle \theta_i, \textbf{1}\rangle)\log(1-\langle \theta_i, \textbf{1}\rangle)$ and $\textbf{1}$ is a vector with all components equal to 1.\\
	\textup{(ii): }We denote the optima of the problems (\ref{eqn:primal-problem}) and (\ref{eqn:dual-problem}) by $\w^*(\alpha, \beta)$ and $\theta^*(\alpha, \beta)$, respectively. Then, the KKT conditions are 
	\begin{align}
	&\w^*(\alpha, \beta) = \frac{1}{\alpha\beta}\calS_{\beta}\big(\frac{1}{n}\sum_{i=1}^{n} A_i\circ \theta_i^*(\alpha, \beta)\big) \tag{KKT-1} \label{eqn:KKT-1}\\
	&\tilde{\theta}_i^*(\alpha, \beta) = \nabla \phi_i(-A_i^T\w^*(\alpha, \beta)), \mbox{ where } \tilde{\theta}_i^*(\alpha, \beta) = (\theta_i^*(\alpha, \beta), 1-\langle \theta_i^*(\alpha, \beta), \textbf{1} \rangle),  i = 1,...,n. \tag{KKT-2} \label{eqn:KKT-2}
	\end{align} 
\end{theorem}

If we define an index set $\calF = \big\{j\in [d]: |[\frac{1}{n}\sum_{i=1}^{n}A_i\circ \theta^*_i(\alpha, \beta)]_j| \leq \beta \big\}$ according to the condition (\ref{eqn:KKT-1}), then it implies the rule below 
\begin{align}
j \in \calF \Rightarrow [\w^*(\alpha, \beta)]_j = 0. \tag{R} \label{eqn:rule-R}
\end{align}
We call the $j$-th feature irrelevant if $j \in \calF$. Moreover, suppose we are given a subset $\hcalF$ of $\calF$, then many coefficients of $\w^*(\alpha,\beta)$ can be inferred by Rule (\ref{eqn:rule-R}). Therefore, the corresponding rows in each $A_i$ can be removed from the problem (\ref{eqn:primal-problem}) and the problem size can be significantly reduced. We formalize this idea in Lemma \ref{lemma:scaled-problem}. 
\begin{lemma} \label{lemma:scaled-problem}
	Given the index set $\hcalF\subseteq \calF$, denote $\hcalF^c = [d]\setminus\hcalF$, then the followings hold\\
	\textup{(i): }$[\w^*]_{\hcalF} = 0$ and $[\w^*]_{\hcalF^c}$ solves the following scaled primal problem:
	\begin{align}
	\min_{\hat{\w} \in \R^{|\hcalF^c|}} \hat{P}(\hat{\w};\alpha, \beta) = \frac{1}{n}\sum_{i=1}^{n}\phi_i(-{}_{\hcalF^c}[A_i]^T \hat{\w})+ \beta r(\hat{\w}). \tag{P'} \label{eqn:primal-problem-scaled}
	\end{align}
	\textup{(ii): }The dual problem of (\ref{eqn:primal-problem-scaled}) takes the form of 
	\begin{align}
	&\min_{\hat{\theta}} \hat{D}(\hat{\theta};\alpha, \beta) = \frac{1}{2\alpha\beta}\|\calS_{\beta}\big(\frac{1}{n}\sum_{i=1}^{n} {}_{\hcalF^c}[A_i]\circ \hat{\theta}_i\big)\|_2^2-\frac{1}{n}\sum_{i=1}^{n}H(\hat{\theta}_i), \tag{D'} \label{eqn:dual-problem-scaled}\\
	&\mbox{s.t. } \hat{\theta} = (\hat{\theta}_1,...,\hat{\theta}_n) \in \Delta_{|\calY_1|-1} \times...\times \Delta_{|\calY_n|-1}. \nonumber 
	\end{align}
	\textup{(iii): } $\theta^*(\alpha, \beta)= \hat{\theta}^*(\alpha, \beta)$ and the KKT conditions between (\ref{eqn:primal-problem-scaled}) and (\ref{eqn:dual-problem-scaled}) are 
	\begin{align}
	&\hat{\w}^*(\alpha, \beta) = \frac{1}{\alpha\beta}\calS_{\beta}\big(\frac{1}{n}\sum_{i=1}^{n} {}_{\hcalF^c}[A]_i\circ \hat{\theta}_i^*(\alpha, \beta)\big), \nonumber \\
	&(\hat{\theta}_i^*(\alpha, \beta), 1-\langle \hat{\theta}_i^*(\alpha, \beta), \textbf{1} \rangle) = \nabla \phi_i(-{}_{\hcalF^c}[A]_i^T\hat{\w}^*(\alpha, \beta)), i = 1,...,n. \nonumber 
	\end{align} 
\end{lemma}

However, Rule (\ref{eqn:rule-R}) is not applicable since it requires the knowledge of $\theta^*(\alpha, \beta)$. Fortunately, the same as most of the existing screening methods, by estimating a region $\Theta$ that contains $\theta^*(\alpha,\beta)$, we can relax Rule (\ref{eqn:rule-R}) into an applicable version, i.e.,  
\begin{align}
j \in \hcalF \mbox{ where }\hcalF = \big\{j\in [d]: \max_{\theta = (\theta_1,...,\theta_n) \in \Theta}|[\frac{1}{n}\sum_{i=1}^{n}A_i \circ \theta_i]_j| \leq \beta \big\} \Rightarrow [\w^*]_j = 0.\tag{R'} \label{eqn:rule-R'}
\end{align}
In view of Rule (\ref{eqn:rule-R'}), the development of our screening method can be sketched as follows:\\
\textbf{Step 1: }Derive the estimation $\Theta$ such that  $\theta^*(\alpha, \beta) \in \Theta$. \\
\textbf{Step 2: }Derive the detailed screening rules via solving the optimization problem in Rule (\ref{eqn:rule-R'}).

\section{The Proposed Method}
In this section, we first present an accurate estimation of the dual optimum by carefully studying of the strong convexity and the structure of the dual problem (\ref{eqn:dual-problem}) of sparse CRF (Section \ref{sec:dual-estimation}). Then, in Section \ref{sec:screening-rule}, we develop our dynamic screening rule by solving a convex optimization problem and give the framework of how to integrate our screening rule with the general training algorithms.  
\subsection{Estimate The Dual Optimum}\label{sec:dual-estimation}
We first show that, for a sufficiently large $\beta$, then for any $\alpha>0$, the primal and dual problem (\ref{eqn:primal-problem}) and (\ref{eqn:dual-problem}) admit closed form solutions. The details are presented in the theorem below. 

\begin{theorem}\label{theorem:beta-max} 
	Let $\beta_{\max} = \frac{1}{n} \|\sum_{i=1}^{n} A_i\frac{ \mathbf{1}}{|\calY_i|}\|_{\infty}$, then for all $\alpha>0$ and $\beta \geq \beta_{\max}$, we have
	\begin{align}
	\w^*(\alpha, \beta) = \mathbf{0} \mbox{ and } \theta^*(\alpha,\beta) = \frac{1}{|\calY_i|} \mathbf{1}, i = 1,...,n. \nonumber
	\end{align}
\end{theorem}
Theorem \ref{theorem:beta-max} tells us that we only need to consider the cases when $\beta \in (0, \beta_{\max}]$. 

Now, we turn to derive the dual optimum estimation. As mentioned in Section \ref{sec:introduction}, since our screening rule is dynamic, we need to  trigger it and estimate the dual optimum during the training process. Thus, we need to denote $\hcalF \subseteq \calF$ as the index set of the irrelevant features we identified in the previous triggering of the screening rule and estimate the dual optimum of the corresponding dual problem (\ref{eqn:dual-problem-scaled}). In addition, we denote the feasible region of the dual problem (\ref{eqn:dual-problem}) as $\calC$ for simplicity, i.e., $\calC = \{\theta: \theta = (\theta_1,...,\theta_n) \in \Delta_{|\calY_1|-1} \times...\times \Delta_{|\calY_n|-1} \}$. 

We find that the objective function of the dual problem (\ref{eqn:dual-problem-scaled}) is $\frac{1}{n}$-strongly convex. Rigorously, we have the lemma below.  
\begin{lemma}\label{lemma:strong-convex}
	Let $\alpha>0, \beta> 0$, $\hcalF\subseteq \calF$ and $\hat{\theta}_1, \hat{\theta}_2 \in \calC$, then the following holds:
	\begin{align}
	\hat{D}(\hat{\theta}_2;\alpha, \beta) \geq \hat{D}(\hat{\theta}_1;\alpha, \beta) + \langle \nabla \hat{D}(\hat{\theta}_1;\alpha,\beta), \hat{\theta}_2-\hat{\theta}_1 \rangle + \frac{1}{2n}\|\hat{\theta}_1-\hat{\theta}_2\|_2^2.\nonumber
	\end{align}
\end{lemma}

The strong convexity is a powerful tool to estimate the optimum $\hat{\theta}^*(\alpha,\beta)$ based on any feasible point in the feasible region $\calC$. Before giving our final estimation, we need the lemma below first.  

\begin{lemma}\label{lemma:dual-estimation}
	Let $\hcalF\subseteq \calF$. For any $\hat{\theta} \in \calC$, we denote $I_1(\hat{\theta}) = n \|\nabla \hat{D}(\hat{\theta};\alpha,\beta)\|_2^2 -2 \hat{D}(\hat{\theta};\alpha,\beta)+ 2 \hat{D}(\hat{\theta}^*(\alpha, \beta);\alpha,\beta)$ and $I_2(\hat{\theta}) = \hat{D}(\hat{\theta};\alpha,\beta)-\hat{D}(\hat{\theta}^*(\alpha, \beta);\alpha,\beta)$, then the followings hold:\\
	\textup{(i):} $I_1(\hat{\theta}) \geq 0$ and $ I_2(\hat{\theta})\geq 0$. \\
	\textup{(ii): } $\lim\limits_{\hat{\theta} \rightarrow \hat{\theta}^*(\alpha, \beta)} I_1(\hat{\theta}) = 0$ and $\lim\limits_{\hat{\theta} \rightarrow \hat{\theta}^*(\alpha, \beta)} I_2(\hat{\theta}) = 0$.\\
	\textup{(iii):} $\|\hat{\theta}^*(\alpha, \beta)- \big(\hat{\theta}_1-n\nabla \hat{D}(\hat{\theta}_1;\alpha,\beta)
	\big)\|^2 \leq  n I_1(\hat{\theta}_1) +2n I_2(\hat{\theta}_2)$ holds for any $\hat{\theta}_1 \in \calC$ and $\hat{\theta}_2\in \calC$.
\end{lemma}

We notice that the item $nI_1(\hat{\theta}_1)+2nI_2(\hat{\theta}_2)$ ($=n^2 \|\nabla \hat{D}(\hat{\theta}_1;\alpha,\beta)\|_2^2 -2n \hat{D}(\hat{\theta}_1;\alpha,\beta)+2n \hat{D}(\hat{\theta}_2;\alpha,\beta)$) in part (iii) of Lemma \ref{lemma:dual-estimation} does not depend on $\hat{\theta}^*(\alpha,\beta)$. Therefore, it is an estimation for $\hat{\theta}^*(\alpha,\beta)$. Lemma \ref{lemma:dual-estimation} also implies that both of $I_1$ and $I_2$ would decrease to 0 when $\hat{\theta}_1$ and $\hat{\theta}_2$ get closer to the optimum $\hat{\theta}^*(\alpha, \beta)$, which makes the estimation more and more accurate. Our final dual estimation is presented in the theorem below.

\begin{theorem}\label{theorem:dual-estimation}
	Let $\alpha>0$, $\beta_{\max}\geq \beta>0$, $\hcalF\subseteq \calF$, then for any $\hat{\theta}\in \calC$, we have
	\begin{align}
	\hat{\theta}^*(\alpha,\beta) \in \Theta = \calB \cap \calH_1 \cap \calH_2\cap...\cap \calH_n,\nonumber 
	\end{align}
	where $\calB = \big \{\theta: \|\theta- \big(\hat{\theta}-n\nabla D(\hat{\theta};\alpha,\beta)
	\big)\|^2 \leq  n I_1(\hat{\theta}) +2n I_2(\hat{\theta}) \big\}$ and $\calH_i = \big\{\theta=(\theta_1,...,\theta_n): \langle\theta_i, \mathbf{1} \rangle \leq 1\}$ with $i=1,...,n$.	
\end{theorem}

Theorem \ref{theorem:dual-estimation} shows that , $\hat{\theta}^*(\alpha,\beta)$ lies in a region $\Theta$, which is the intersection of a ball and $n$ half spaces. From Lemma \ref{lemma:dual-estimation}, we can see that the radius of the ball $\calB$ would decrease to 0. Hence, our estimation $\Theta$ would become more and more accurate as the training process going on.

\textbf{Discussion.} From Theorem \ref{theorem:dual-estimation}, we can see that our dual estimation does not depend on the duality gap. In addition, the vector $
\hat{\theta}$ in our estimation can be any point in the feasible region $\calC$, hence, our estimation can be updated during the training process. We also notice that the volume of our estimation $\Theta$ can be reduced to 0 when $\hat{\theta}$ converges to optimum. Therefore, our estimation can be used to develop dynamic screening rules.
\subsection{The Proposed Screening Rule}\label{sec:screening-rule}
Based on our dual estimation, we can develop the detailed screening rules by solving the optimization problem below:
\begin{align}
s_j =\max_{\theta=(\theta_1,...,\theta_n) \in \Theta} |[\frac{1}{n}\sum_{i=1}^{n}A_i \circ \theta_i]_j|. j = 1...,d. \label{eqn:screening-rule-optimization}
\end{align} 
Due to the complex structure of $\Theta$, directly solving the problem above is time consuming. Denote $\Theta_k = \calB \cap \calH_k$, $k=1,...,n$, We relax problem (\ref{eqn:screening-rule-optimization}) into:
\begin{align}
s_j =&\max_{\theta=(\theta_1,...,\theta_n) \in \Theta} |[\frac{1}{n}\sum_{i=1}^{n}A_i \circ \theta_i]_j| \nonumber \\
\leq& \min \Big \{  s_{j,k} = \max_{\theta=(\theta_1,...,\theta_n) \in \Theta_k} |[\frac{1}{n}\sum_{i=1}^{n}A_i \circ \theta_i]_j|, k =1,...,n \Big \}=\hat{s}_j, j = 1...,d, \label{eqn:screening-rule-optimization-relaxed}
\end{align} 
The value of $s_{j,k}$ can be calculated by solving the following two problems:
\begin{align}
\begin{cases}
s_{j,k}^+ = -\min_{\theta=(\theta_1,...,\theta_n) \in \Theta_k} -[\frac{1}{n}\sum_{i=1}^{n}A_i \circ \theta_i]_j,\\
s_{j,k}^- = -\min_{\theta=(\theta_1,...,\theta_n) \in \Theta_k} [\frac{1}{n}\sum_{i=1}^{n}A_i\circ  \theta_i]_j,
\end{cases} 
j = 1...,d. \label{eqn:two-problems}
\end{align}  
Clearly, we have $s_{j,k} = \max \big \{s_{j,k}^+, s_{j,k}^-\big \}$. We notice that  we can written the item $A_i \circ \theta_i$ as:
\begin{align}
A_i \circ \theta_i &= [A_i]_{:,|\calY_i|-1}\theta_i +(1-\langle \theta_i, \textbf{1}\rangle[A_i]_{:, |\calY_i|}) \nonumber \\
&=\big([A_i]_{:,|\calY_i|-1}-[A_i]_{:, |\calY_i|}\textbf{1}^T\big) \theta_i+[A_i]_{:, |\calY_i|} \mbox{ (the item } [A_i]_{:, |\calY_i|}\textbf{1}^T \mbox{ here is a matrix)}\nonumber  
\end{align}
Thus , the two problems in (\ref{eqn:two-problems}) can be written uniformly as 
\begin{align}
\min_{\a} \langle \b, \a \rangle, s.t.  \langle \p, \a \rangle \leq c  \mbox{ and } \|\a- \a_0 \|_2^2 \leq r^2. \label{eqn:convex-problem}
\end{align}
Using the standard Lagrangian multiplier method, we can obtain a closed form solution for Problem (\ref{eqn:convex-problem}). Rigorously, we have the Theorem below.

\begin{theorem}\label{theorem:closed-form-solution}
	Let $d = \frac{\langle \p, \a_0 \rangle - c}{r \| \p \|} $, the closed form solution of this minimization problem can be calculated as
	\begin{enumerate}
		\item[\textup{i.}] When $b\in (-1,1]$, if $\langle \p, \b \rangle \neq \pm \| \p \| \| \b \|$ and 
		\begin{enumerate}
			\item[\textup{1.}] if $\langle \p,\b \rangle \geq  d \| \p\| \|\b\|$, then the optimal value is $\langle \b, \a_0\rangle - r \|\b\|$.
			\item[\textup{2.}] if $\langle \p,\b \rangle < d \| \p\| \|\b\|$, defining $\Delta := 4(1-b^2)\|\p\|^4 b^2 (\|\p\|^2\|\b\|^2 - \langle \p,\b \rangle^2) \geq 0$, then the optimal value is 
			\[
			-\frac{1}{4\lambda_1} \|\b+\lambda_2 \p \|^2 + \langle \b+\lambda_2 \p, \alpha_0 \rangle - (\lambda_1 r^2 + \lambda_2 c),
			\] 
			where we define that
			\[
			\lambda_1 = \frac{\|\b + \lambda_2 \p \|}{2r},
			\]
			\[
			\lambda_2 = \frac{-2(1-b^2) \| \p \|^2 \langle\p,\b \rangle + \sqrt{\Delta}}{2\|\p\|^4 (1-b^2)}
			\]
			\item[\textup{3.}] if $\langle \p,\b \rangle < b \| \p\| \|\b\|$ and $b\in (-1,1)$,
		\end{enumerate}
		
		\item[\textup{ii.}] When $b\in (-1,1]$, if $\langle \p, \b \rangle =  \| \p \| \| \b \|$, the optimal value is $\langle \b, \a_0\rangle - r \|\b\|$.
		
		\item[\textup{iii.}] When $b\in (-1,1]$, if $\langle \p, \b \rangle = - \| \p \| \| \b \|$, the optimal value is $\frac{c\|\b\|}{\|\p\|}$.
		
		\item[\textup{iv.}] When $d \in (-\infty, -1]$, which means the feasible set is equal to $\{ \a : \| \a - \a_0 \|^2 \leq r^2\}$ , then the optimal value is $\langle \b, \a_0\rangle - r \|\b\|$.
		
		\item[\textup{v.}] When $d \in (1, +\infty)$, which means the feasible set is empty, then the optimal value is $\emptyset$.
	\end{enumerate}
	
\end{theorem}
Therefore, we can solve Problem (\ref{eqn:convex-problem}) efficiently by Theorem \ref{theorem:closed-form-solution}. We are now ready to present the final screening rule. 

\begin{theorem}\label{theorem:screeing-rule}
	Given the dual estimation $\Theta$, then the followings hold:\\
	\textup{(i): }The feature screening rule takes the form of 
	\begin{align}
	\hat{s}_j \leq \beta \Rightarrow [\w^*(\alpha, \beta)]_i = 0, \forall i \in \hcalF^c. \tag{R$^*$} \label{eqn:screening-rule-final}
	\end{align}  
	\textup{(ii): } If new irrelevant features are identified, we can update the index set $\hcalF$ by:
	\begin{align}
	\hcalF \leftarrow \hcalF \cup \Delta \hcalF \mbox{ with }\Delta \hcalF =\big \{j: \hat{s}_j \leq \beta, j \in \hcalF^c \big\}. \label{eqn:F-update}
	\end{align} 
\end{theorem}
Below, we present a framework of how to integrate our screening rules in a dynamic manner with the general training algorithms for sparse CRFs.  $\hat{G}(\hat{\w},\hat{\theta})$ in step 6 is the duality gap, i.e.,  $\hat{G}(\hat{\w},\hat{\theta})= \hat{P}(\hat{\w};\alpha, \beta)+\hat{D}(\hat{\theta};\alpha, \beta)$. 

\begin{algorithm}[htb]\caption{Safe Screening for Sparse CRF}
	\begin{algorithmic}[1]
		\STATE {\bfseries Input: } An optimization algorithm $\calA$ for problem (\ref{eqn:primal-problem}), $\alpha>0, \beta>0, \epsilon > 0, \hat{\w}_0$. 
		\STATE {\bfseries Initialize:} $\hcalF =\emptyset, \hat{\w} = \hat{\w}_0, 0<\gamma<1$, $g = \infty$.
		\REPEAT
		\STATE Run $\calA$ on problem (\ref{eqn:primal-problem-scaled}) to update $\hat{\w}$. 
		\STATE Update $\hat{\theta}$ by $\hat{\theta}_i = \nabla \phi(-{}_{\hcalF^c}[A_i]^T\hat{\w}), i = 1,...,n$. 
		\IF{ dual gap $\hat{G}(\hat{\w},\hat{\theta})<\gamma g$}
		\STATE Run the screening rule (\ref{eqn:screening-rule-final}). 
		\STATE Update the irrelevant feature set $\hcalF$ by (\ref{eqn:F-update}).
		\IF{$\Delta\hcalF\neq \emptyset$}
		\STATE Update (\ref{eqn:primal-problem-scaled}) and (\ref{eqn:dual-problem-scaled}) according to $\hcalF$ and update $\hat{\w}$ and $g$ by:
		\begin{align}
		\hat{\w} \leftarrow [\hat{\w}]_{\hcalF^c} \mbox{ and } g\leftarrow \hat{G}(\hat{\w},\hat{\theta}) \mbox{ with } \hat{\theta}_i = \nabla \phi(-{}_{\hcalF^c}[A_i]^T\hat{\w}), i = 1,...,n.\nonumber
		\end{align}
		\ELSE
		\STATE Update $g$ by:
		\begin{align}
		g\leftarrow \hat{G}(\hat{\w},\hat{\theta}) \mbox{ with } \hat{\theta}_i = \nabla \phi(-{}_{\hcalF^c}[A_i]^T\hat{\w}), i = 1,...,n.\nonumber
		\end{align}
		
		\ENDIF
		\ENDIF
		\UNTIL {$G(\hat{\w}, \hat{\theta}) < \epsilon$.}
		\STATE {\bfseries Return:} $\w \in \R^d$ with $[\w]_{\hcalF}=0$ and $[\w]_{\hcalF^c}=\hat{\w}$. 
	\end{algorithmic}\label{alg:screening}
\end{algorithm}

In real application, we usually need to solve problem (\ref{eqn:primal-problem}) at a sequence of parameter values $\{(\alpha_i, \beta_i), i = 1,...,K\}$ to find the optimal values of $\alpha, \beta$. In this case, we can use the optimal solution $\w^*(\alpha_{i-1}, \beta_{i-1})$ as the warm starter of the problem (\ref{eqn:primal-problem}) at $(\alpha_i, \beta_i)$. 

\begin{remark}
	Recall that $\hat{s}_j = \min\{ s_{j,k}, k =1,...,n  \}$. Therefore, to calculate $\hat{s}_j$, we need to solve $2n$ problems in the form of  (\ref{eqn:convex-problem}), which would be time consuming. In real applications, we can randomly choose $k\in [n]$ each time we triggering our screening rule and relax $\hat{s}_j$ to $s_{j,k}$.
\end{remark}

\begin{remark}
	The parameter $\gamma$ in Algorithm \ref{alg:screening} controls the frequency how often we trigger the rule \ref{eqn:screening-rule-final}. Theoretically, it can be any number lying in (0,1). The larger value, the higher frequency to trigger \ref{eqn:screening-rule-final} and more computational time consumed by it.
\end{remark}

\section{Experiments}\label{sec:experiments}
We evaluate our screening method through numerical experiments on both synthetic and real data sets in terms of three measurements. The first one is the rejection ratio over iterations: $\frac{d_i}{d^*}$, where $d^*$ is the number of the zero valued coefficients in the final learned model and $d_i$ is number of the irrelevant features identified by the screening rule after the $i$-th iteration. The second measurement in speedup, that is the ratio of the running times of the training algorithm with our screening and with screening. The last one is the reservation ratio over iterations: $\frac{d-d_i}{d}$ with $d$ is the feature dimension. Actually, it is the ratio of the problem size after $i$-th iteration to the original problem size. The accuracy $\epsilon$ and the parameter $\gamma$ in Algorithm \ref{alg:screening} are set to be $10^{-6}$ and $0.5$, respectively. 

For each dataset, we solve problem (\ref{eqn:primal-problem}) at a sequence of turning parameter values. Specially, we fix $\alpha =1$ and compute the $\beta_{\max}$ by Theorem \ref{theorem:beta-max}. Then, we select 100 values of $\beta$ that are equally spaced on the logarithmic scale of $\beta/\beta_{\max}$ from 1 to 0.1. Thus, we solve problem (\ref{eqn:primal-problem}) at 100 pairs of parameter values on each dataset. We write the code in Matlab and perform all the computations on a single core of Intel(R) Core(TM) i7-5930K 3.50GHz, 32GB MEM. 

\subsection{Experiments on Synthetic Datasets}
We evaluate our screening rules on 3 datasets named syn1, syn2 and syn3 and their sample and feature size $(n,d)$ are $(10000,1000), (10000,10000)$ and $(1000,10000)$, respectively. Each dataset has $C=10$ classes and each class has $n/C$ samples. We write each point $\x$ as $[\x_1;\x_2]$, where $\x_1 = [\x_1^1;...;\x_1^C] \in \R^{0.02d}$ with $\x_1^k \in \R^{0.02p/C}$ and $\x_2 \in \R^{0.98d}$. If $\x$ belongs to the $k$-th class, then we sample $\x_1^k$ form a Gaussian distribution $\calG = N(\u, 0.75 \mathbf{I})$ with $\u = 1.5 \mathbf{1}, \mathbf{I} \in \R^{(0.02p/C)\times(0.02p/C)}$ and we sample other components in $\x_1$ from the standard Gaussian distribution $N(0,1)$. Each coefficient in $\x_2$ would be sampled from distribution $N(0,1)$ with probability $\eta = 0.2$ or 0 with probability $1-\eta$. The feature vector $F(\x, y)\in \R^{dC}$ is defined as:
\begin{align}
[F(\x,y)]_j = [\x]_i I(y=c), i=1,...,d, c=1,...,C, \nonumber
\end{align} 
where $j=i+(c-1)d$ and $I(\cdot)$ is the index function.

\begin{figure*}[htb!]
	\begin{center}
		\subfigure[$\beta/\beta_{\max}$=0.1]{\includegraphics[scale=0.23]{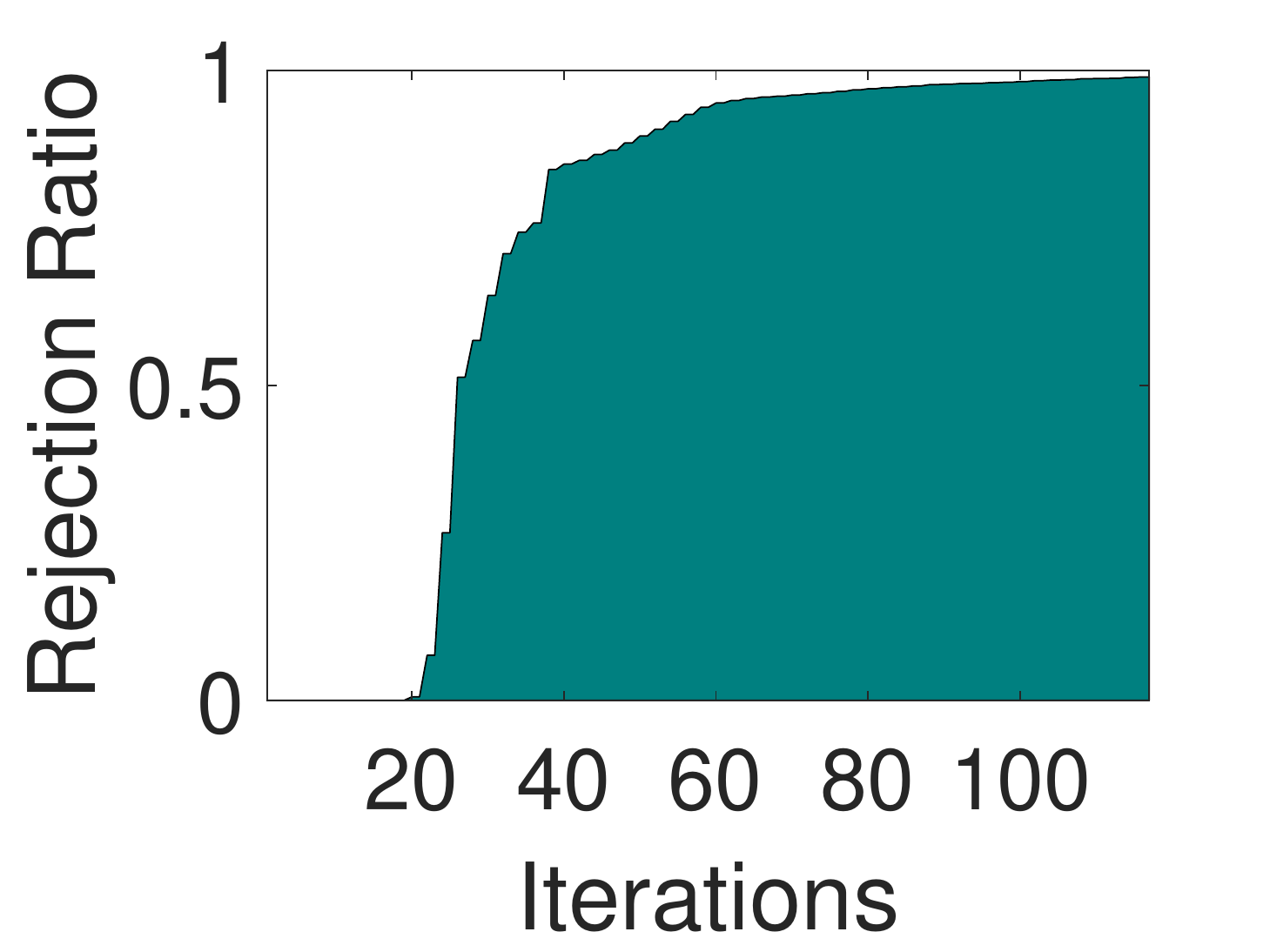}}
		\subfigure[$\beta/\beta_{\max}$=0.2]{\includegraphics[scale=0.23]{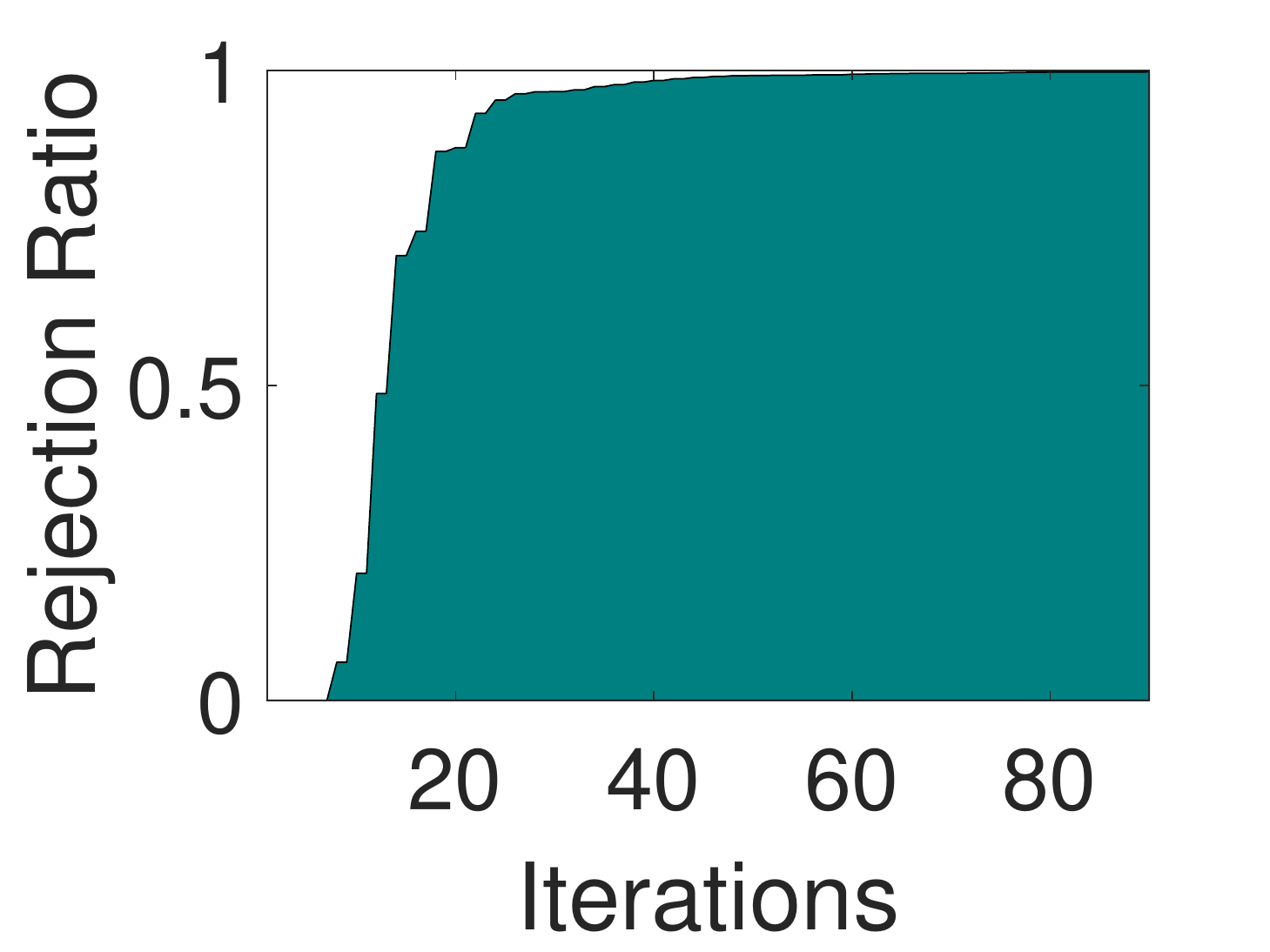}}
		\subfigure[$\beta/\beta_{\max}$=0.5]{\includegraphics[scale=0.23]{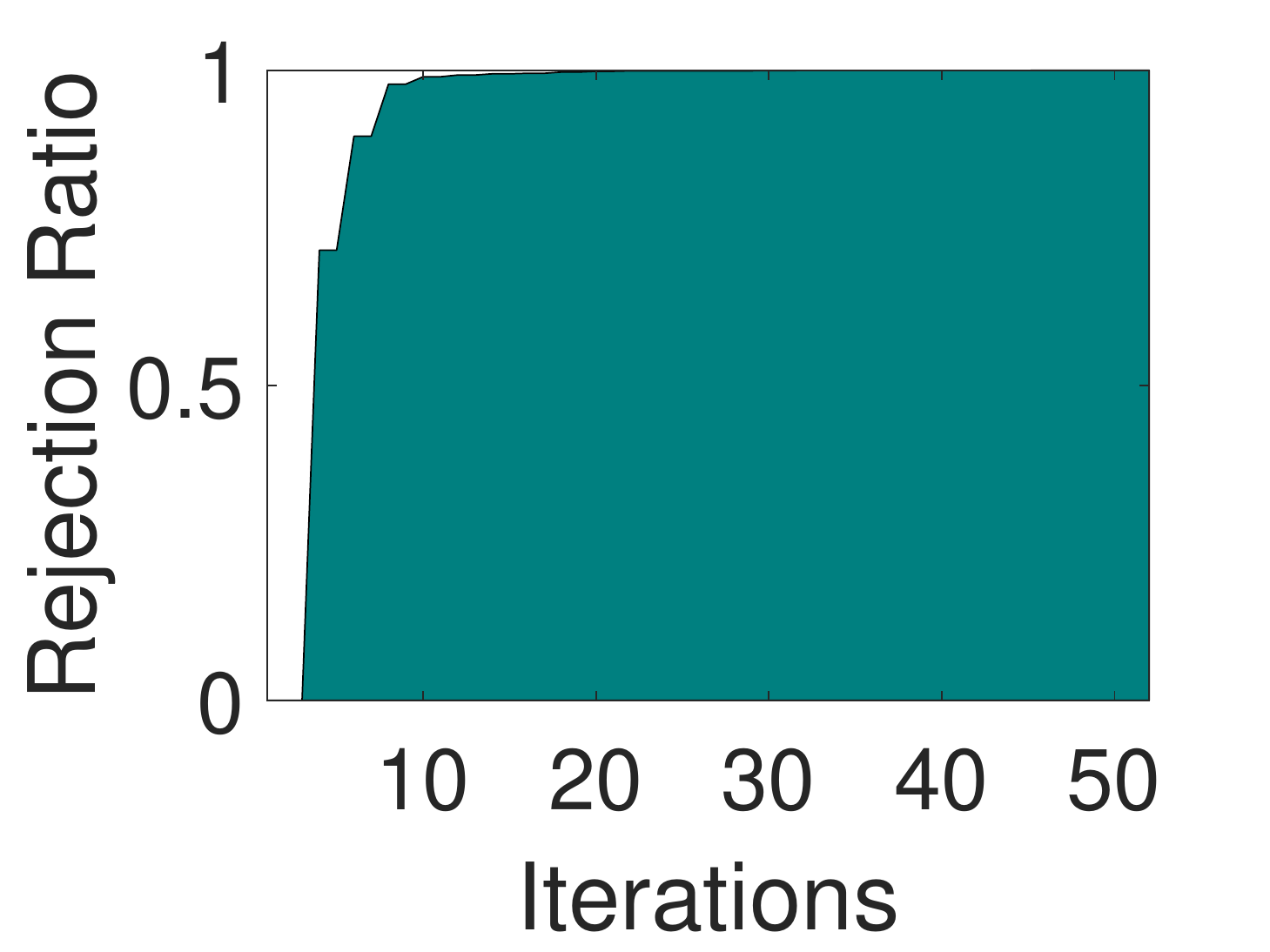}}
		\subfigure[$\beta/\beta_{\max}$=0.9]{\includegraphics[scale=0.23]{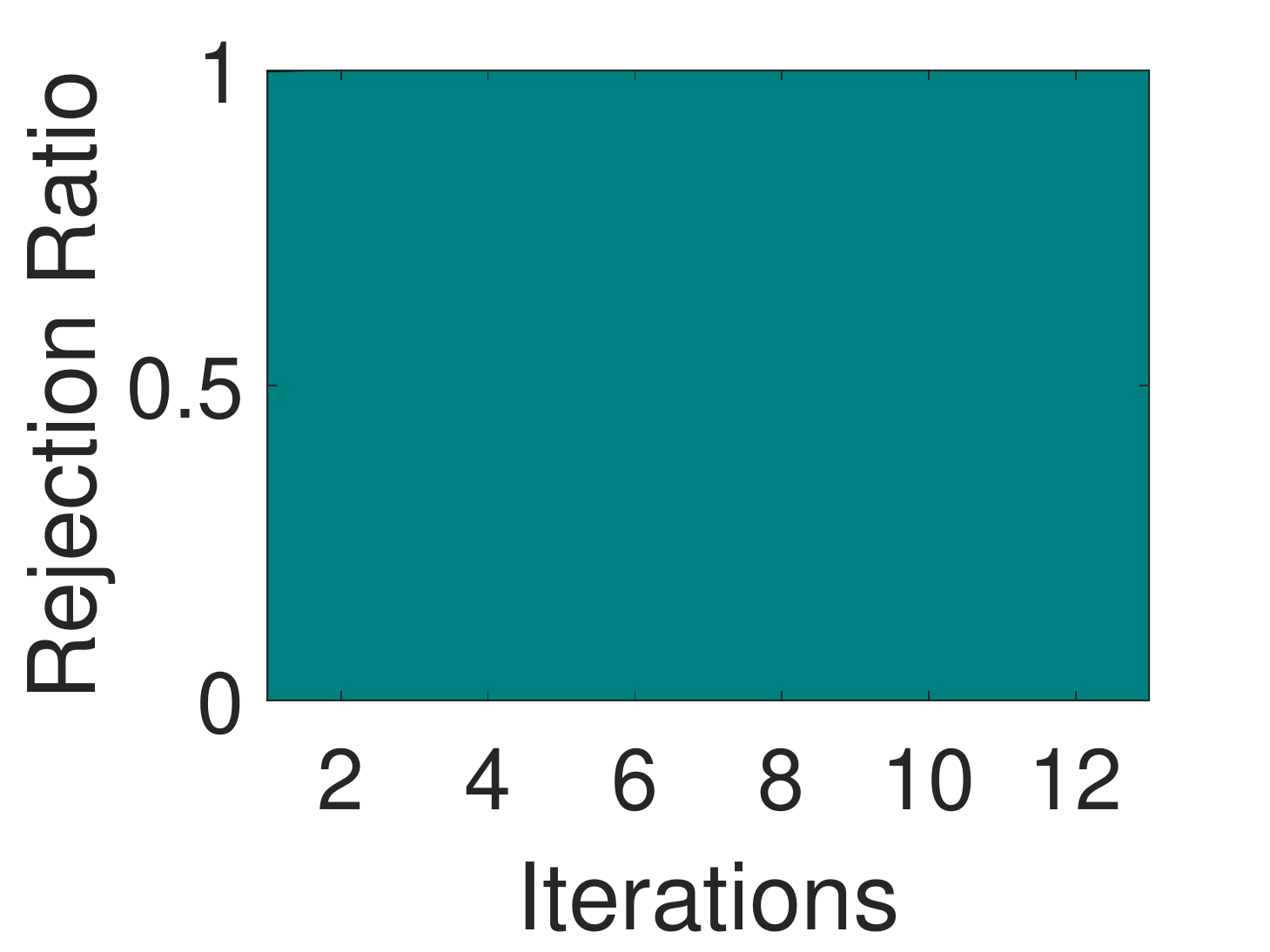}}
		\caption{Rejection ratios of our screening method over iterations on the syn3.}
		\label{fig:reject-ratios-syn}
	\end{center}
\end{figure*}

Due to the space limitation, we only report the rejection ratios of our screening method on syn3. Figure \ref{fig:reject-ratios-syn} shows that our approach can identify the irrelevant features incrementally during the training process. It can finally find almost all the irrelevant features in the end of the training process. 

\begin{figure*}[htb!]
	\begin{center}
		\subfigure[$\beta/\beta_{\max}$=0.1]{\includegraphics[scale=0.23]{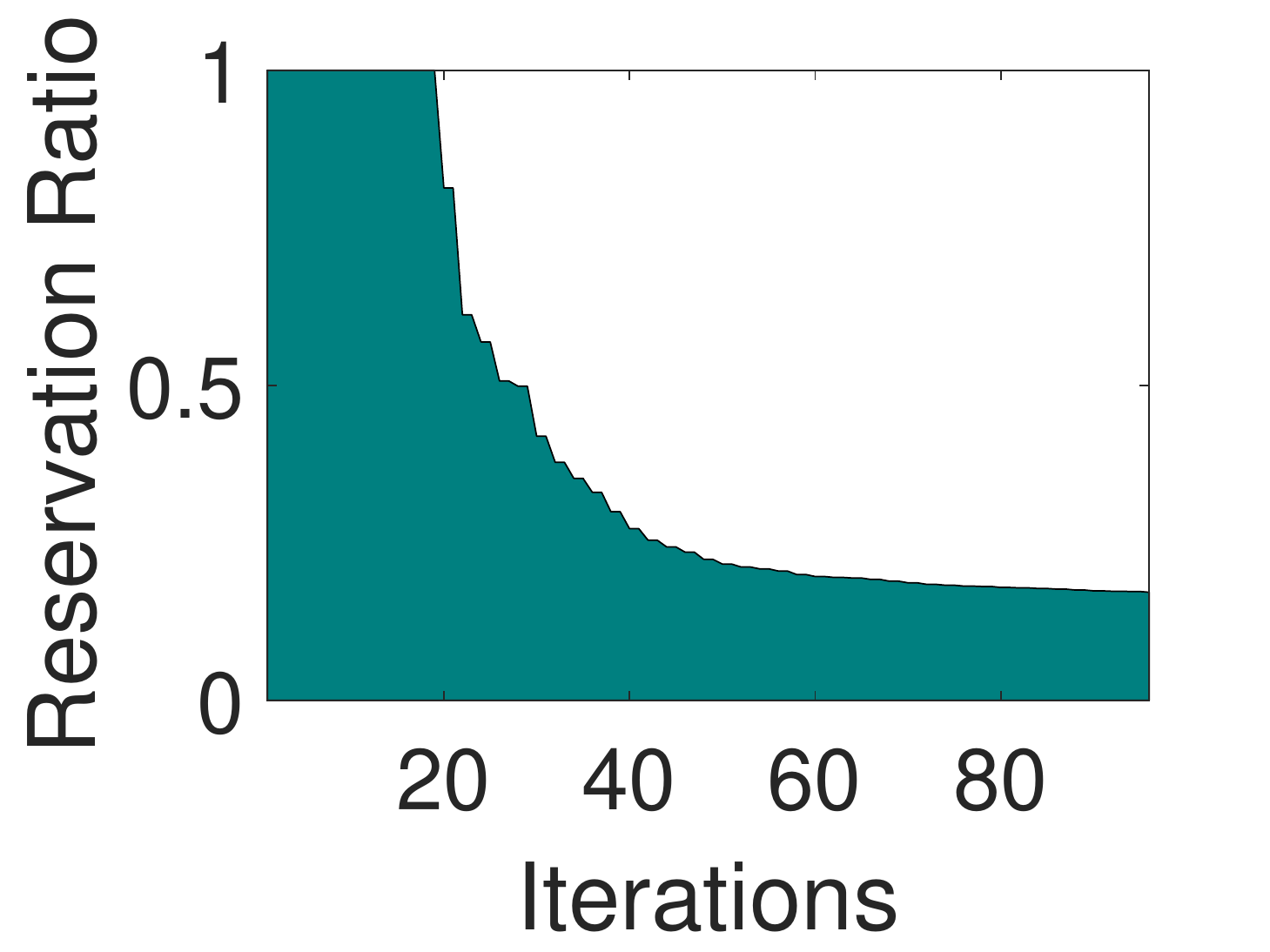}}
		\subfigure[$\beta/\beta_{\max}$=0.2]{\includegraphics[scale=0.23]{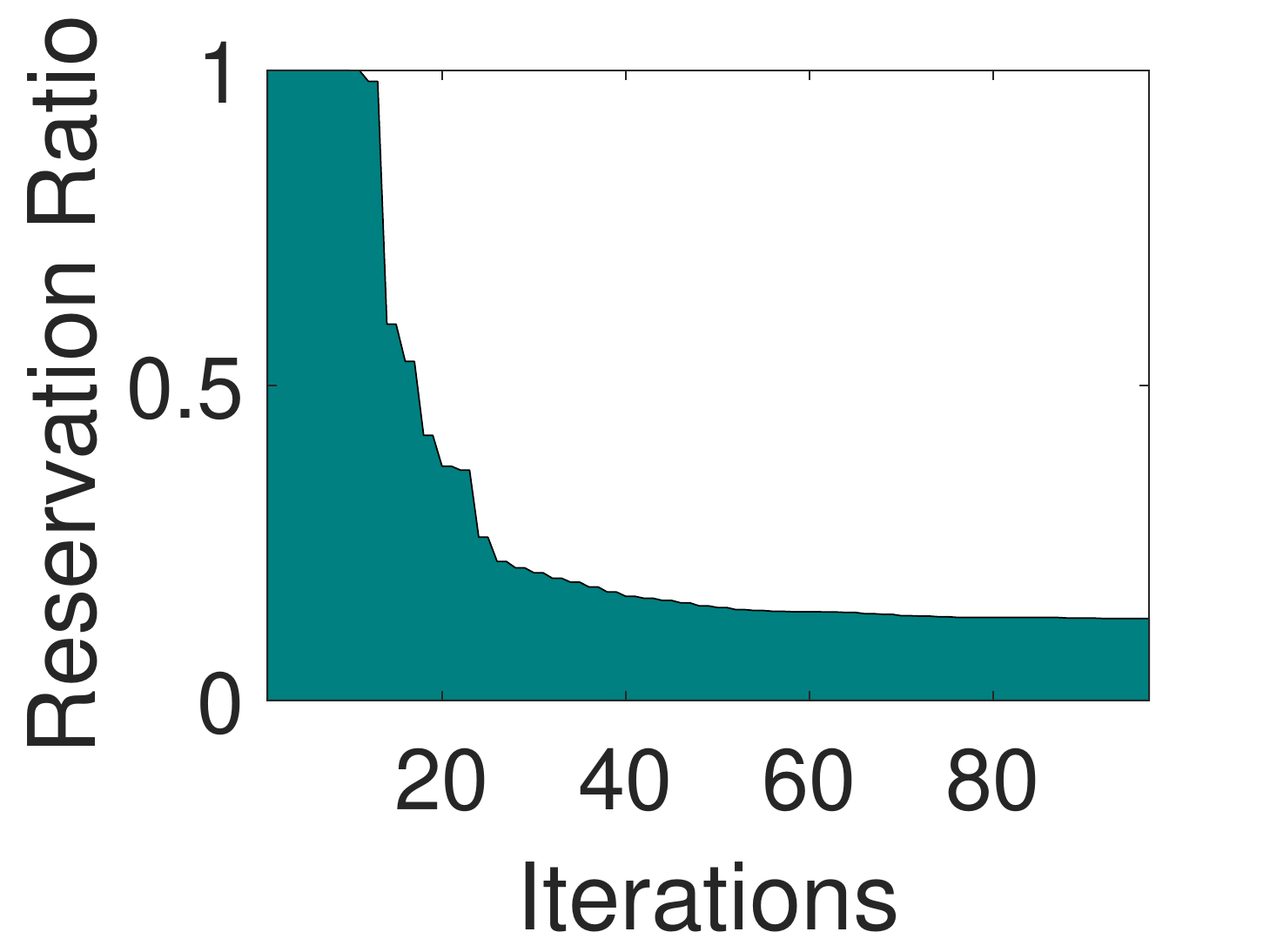}}
		\subfigure[$\beta/\beta_{\max}$=0.5]{\includegraphics[scale=0.23]{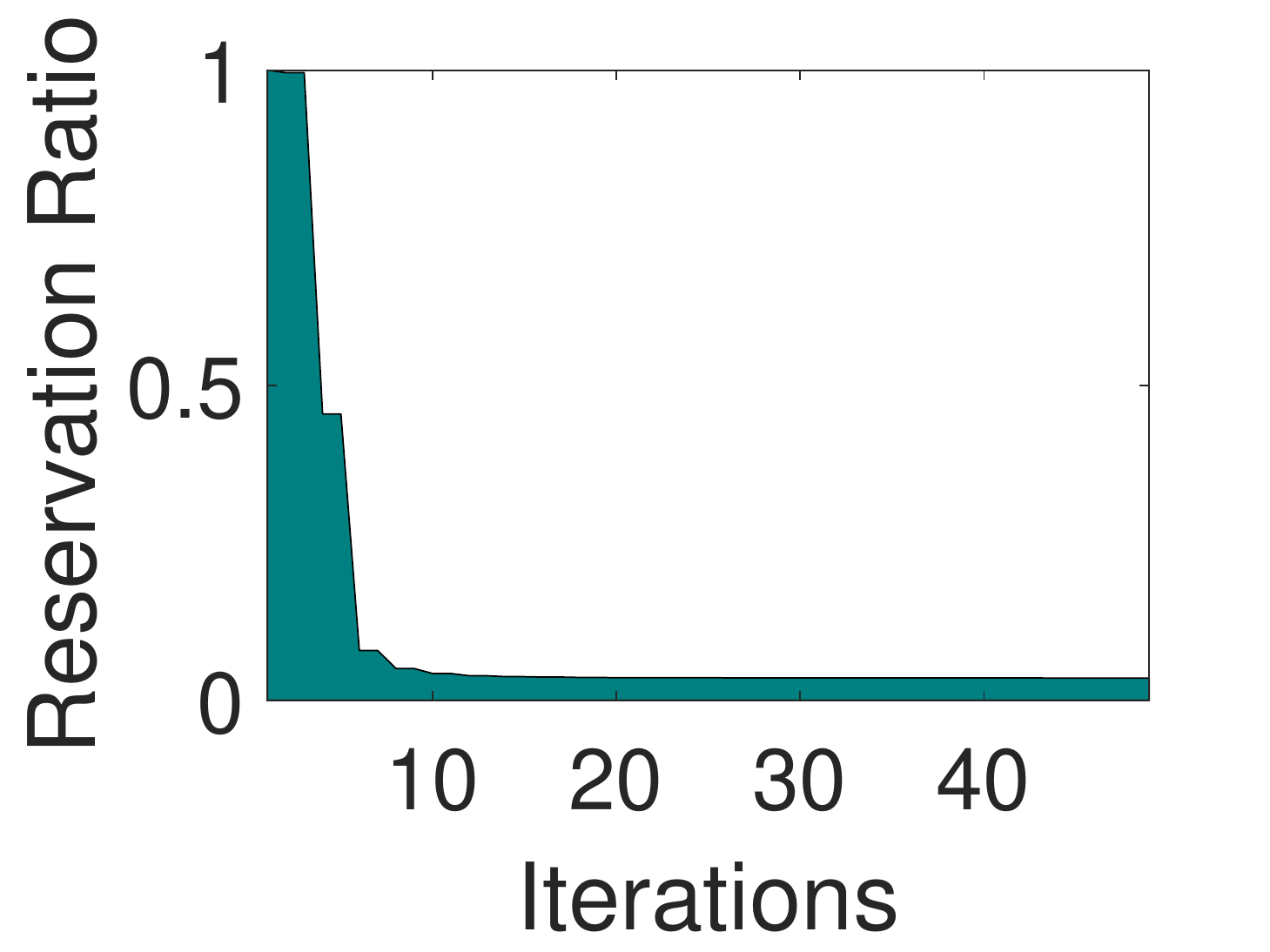}}
		\subfigure[$\beta/\beta_{\max}$=0.9]{\includegraphics[scale=0.23]{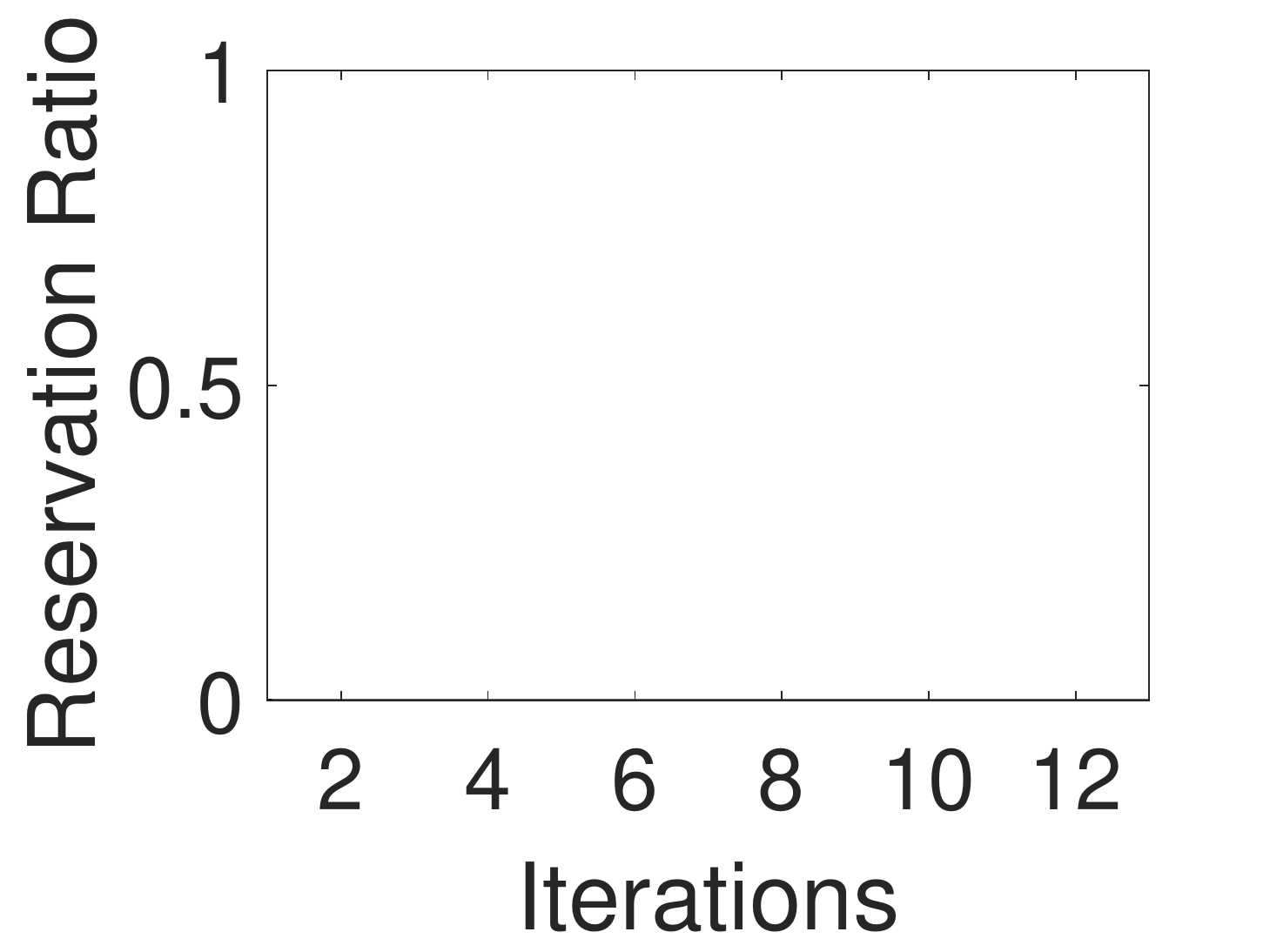}}
		\caption{Reservation ratios of our screening method over iterations on the syn3}
		\label{fig:scaling-ratios-syn}
	\end{center}
\end{figure*}

Figure \ref{fig:scaling-ratios-syn} shows the reservation ratios of our approach over the iterations. We can see that the problem size  can be reduced very quickly over the iterations especially when $\beta$ is large. This reservation ratios imply that we can achieve a significant speedup. 

We report the running times of the training algorithm with and without screening in Table \ref{table:run-time-sync}. It shows that our approach achieves significant speedups, that is up to 5.2 times.  Moreover, we can see that the time cost of our screening method is negligible. 

\begin{table*}[htb!]
	\centering
	\caption{Running time (in seconds) for solving problem (\ref{eqn:primal-problem}) at $100$ pairs of parameter values.}
	\label{table:run-time-sync}
	{\footnotesize
		\begin{tabular}{|c|c|c|c|c|}
			\hline
			\multirow{2}{*}{data} & \multirow{2}{*}{solver} & \multicolumn{3}{c|}{solver+screening} \\ \cline{3-5} 
			&  & screening & solver & speedup \\ \hline
			syn1 &443&12&138&2.95 \\ \hline
			syn2 &1652&16&366&4.30 \\ \hline
			syn3 &3110 &30 & 568& 5.20 \\ \hline
			OCR &6723&55 & 1195& 5.4 \\ \hline
		\end{tabular}
	}
\end{table*}

\subsection{Experiments on Real Datasets}
In this experiment, we evaluate the performance of our screening method on the optical character recognition (OCR) dataset \cite{kassel1995comparison}. OCR has about 50,000 handwritten words, with average length of 8 characters from 150 human subjects. Each word has been divided into characters and each character is presented by a $16\times 8$ gray-scale feature map. Our task in this experiment is handwriting recognition using sparse CRF model. 

\begin{figure*}[htb!]
	\begin{center}
		\subfigure[$\beta/\beta_{\max}$=0.1]{\includegraphics[scale=0.23]{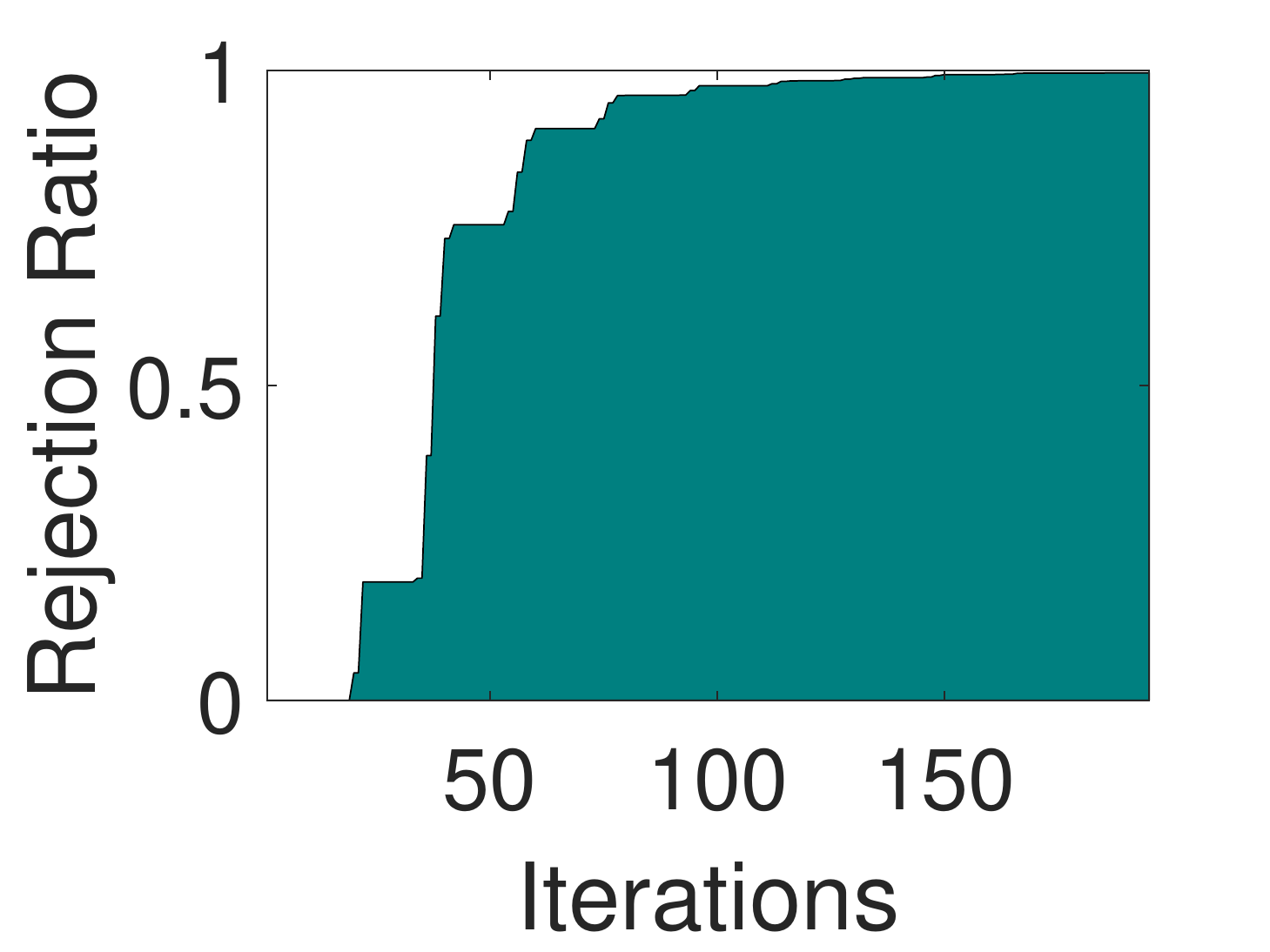}}
		\subfigure[$\beta/\beta_{\max}$=0.2]{\includegraphics[scale=0.23]{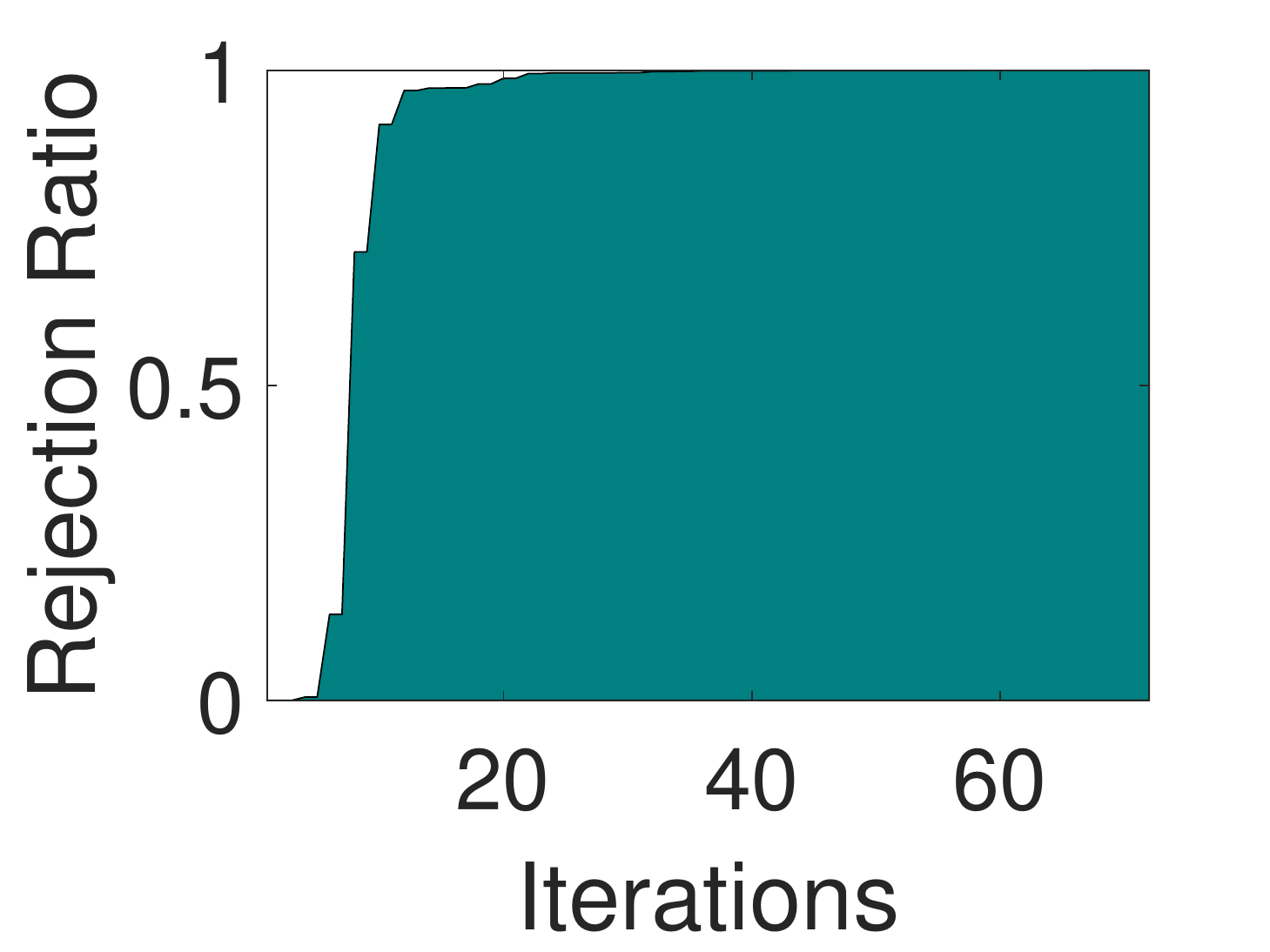}}
		\subfigure[$\beta/\beta_{\max}$=0.5]{\includegraphics[scale=0.23]{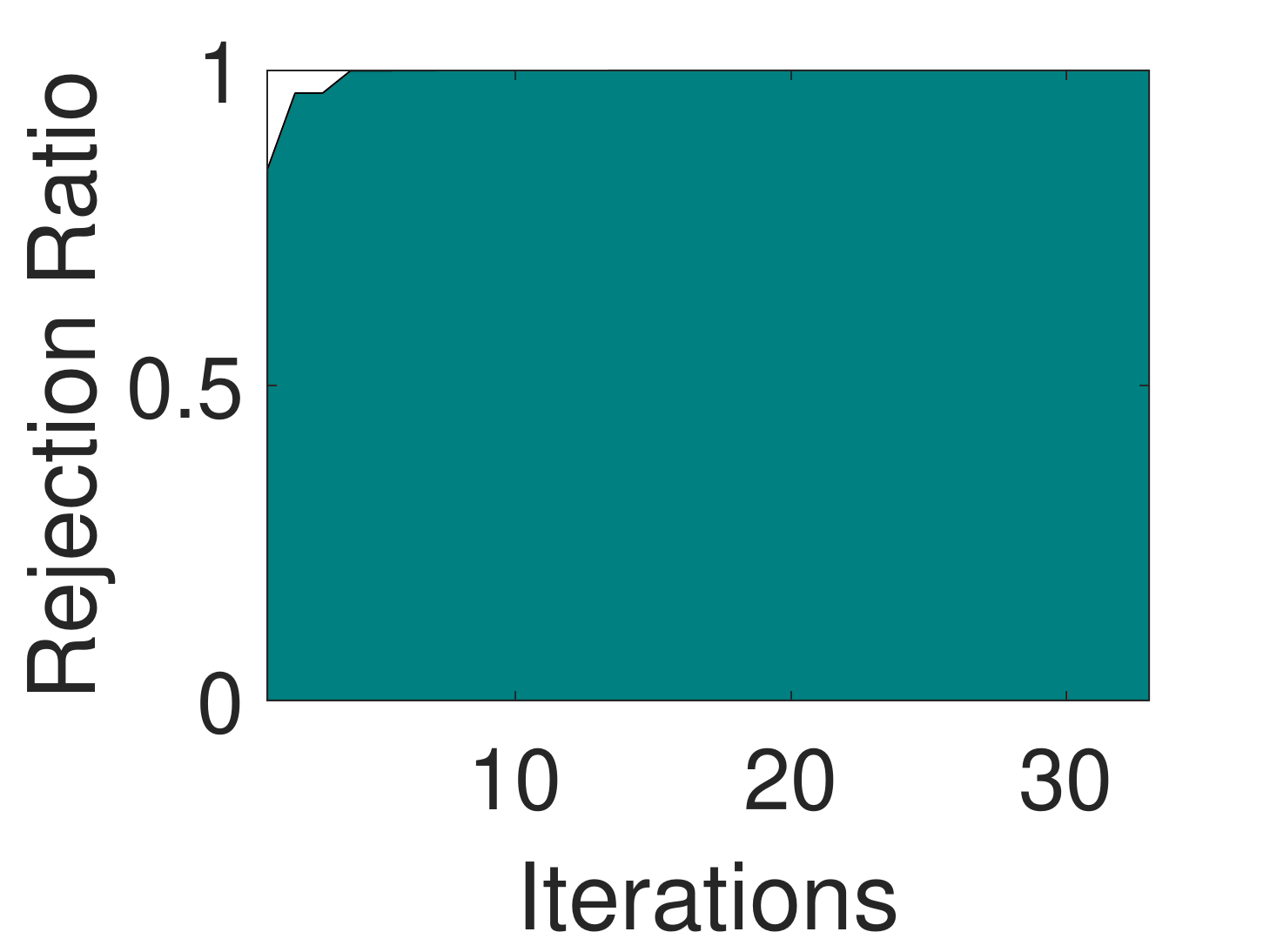}}
		\subfigure[$\beta/\beta_{\max}$=0.9]{\includegraphics[scale=0.23]{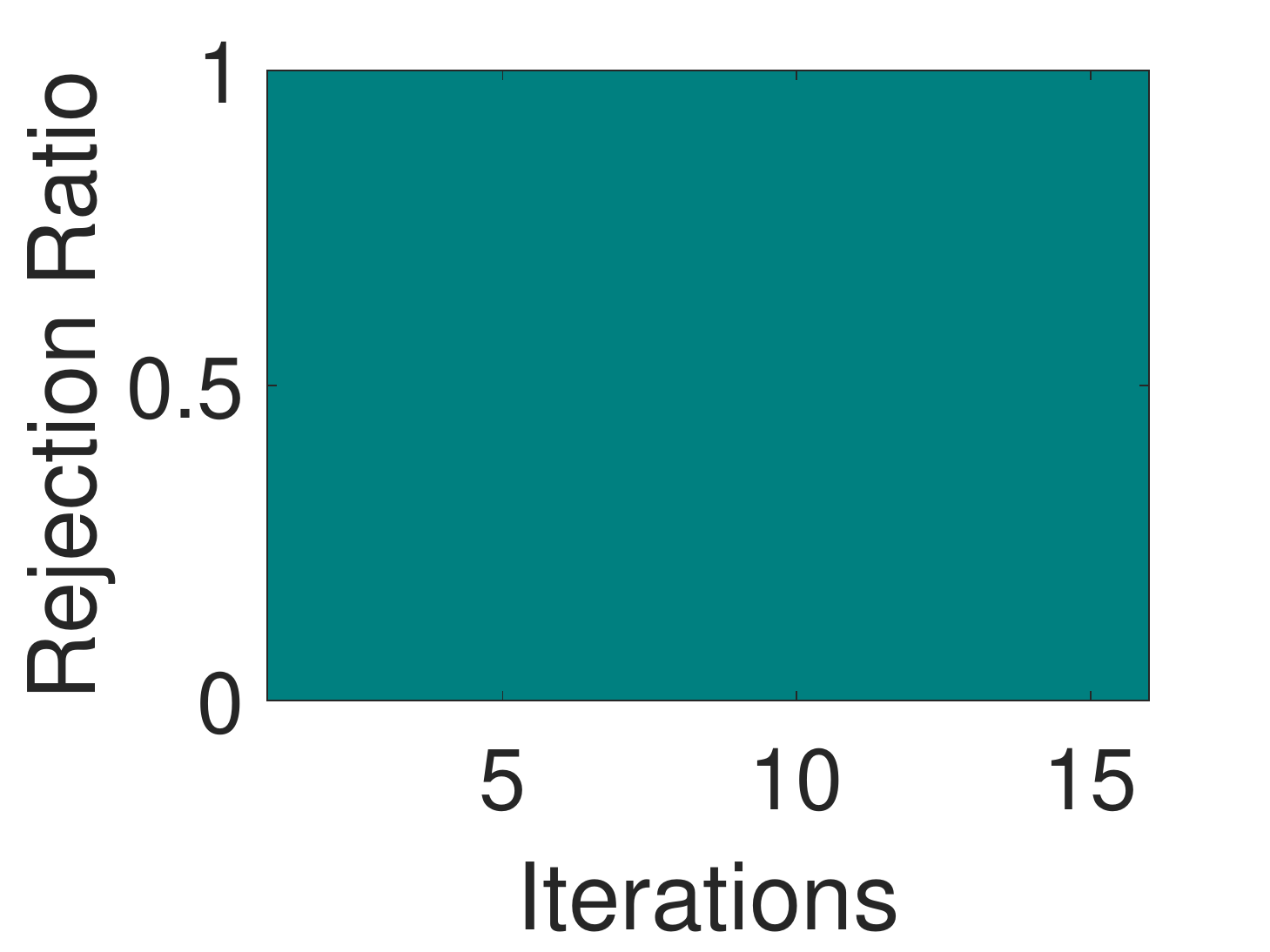}}
		\caption{Rejection ratios of our screening method over the iterations on OCR.}
		\label{fig:reject-ratios-real}
	\end{center}
\end{figure*}
\begin{figure*}[htb!]
	\begin{center}
		\subfigure[$\beta/\beta_{\max}$=0.1]{\includegraphics[scale=0.23]{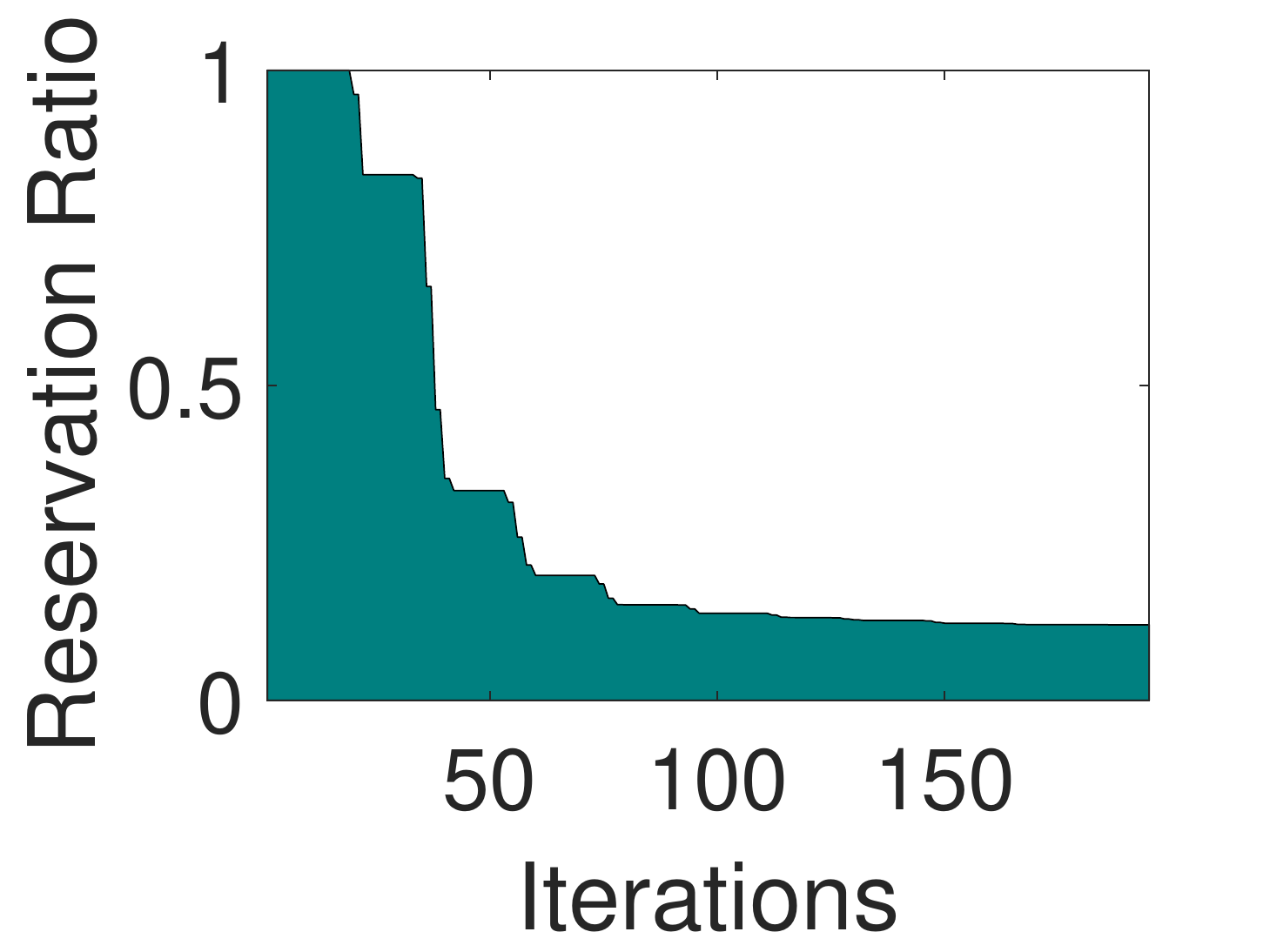}}
		\subfigure[$\beta/\beta_{\max}$=0.2]{\includegraphics[scale=0.23]{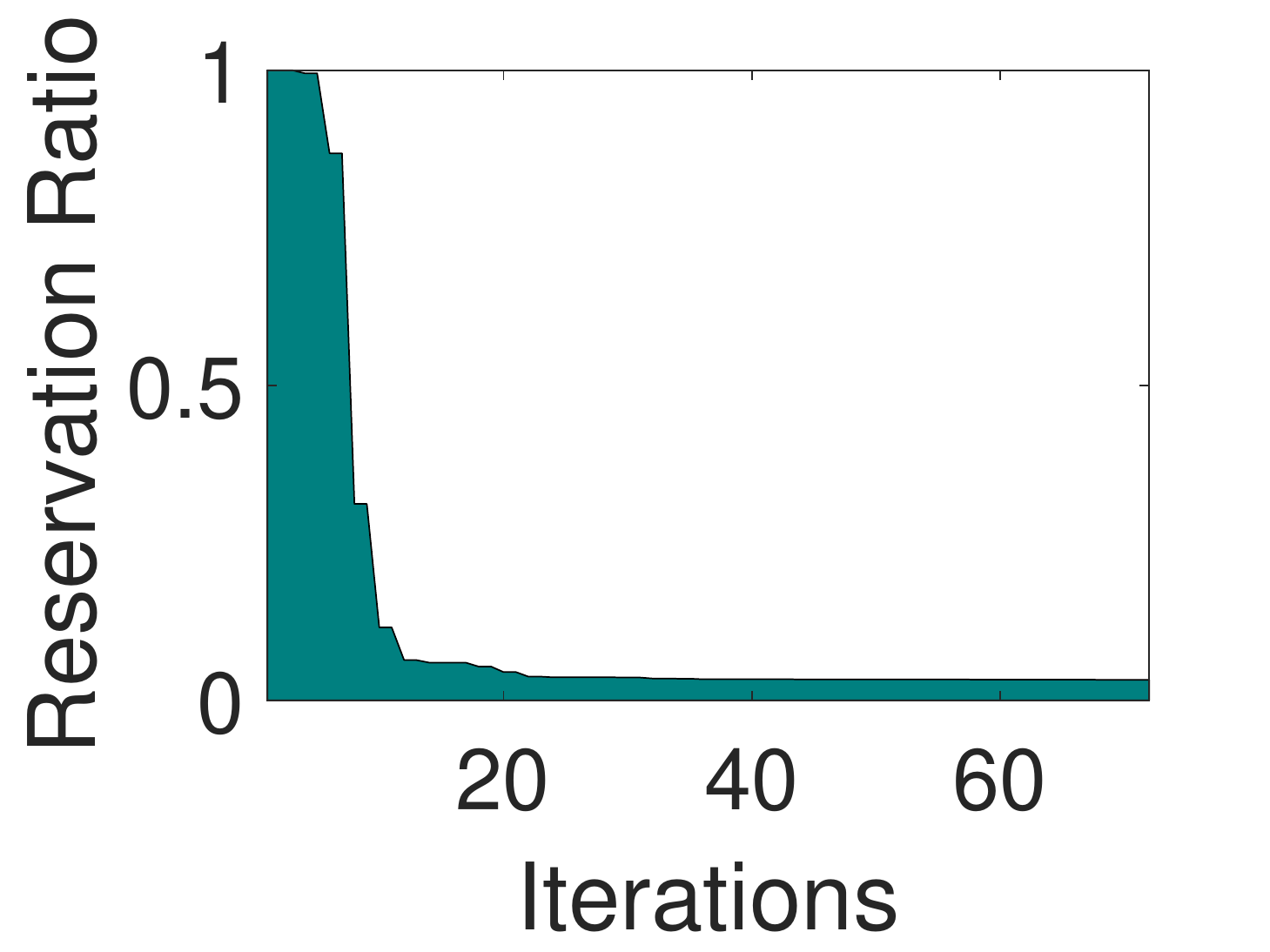}}
		\subfigure[$\beta/\beta_{\max}$=0.5]{\includegraphics[scale=0.23]{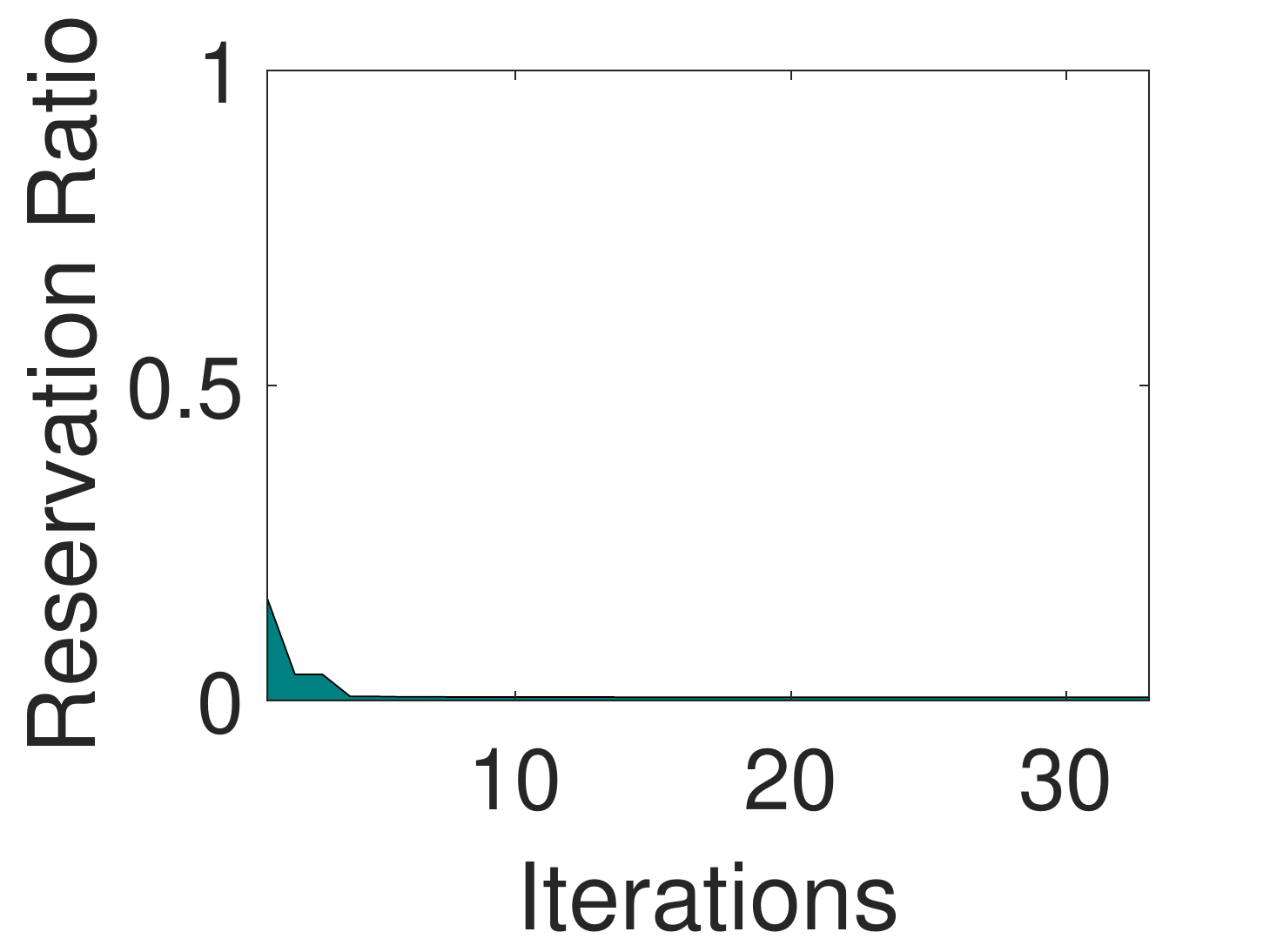}}
		\subfigure[$\beta/\beta_{\max}$=0.9]{\includegraphics[scale=0.23]{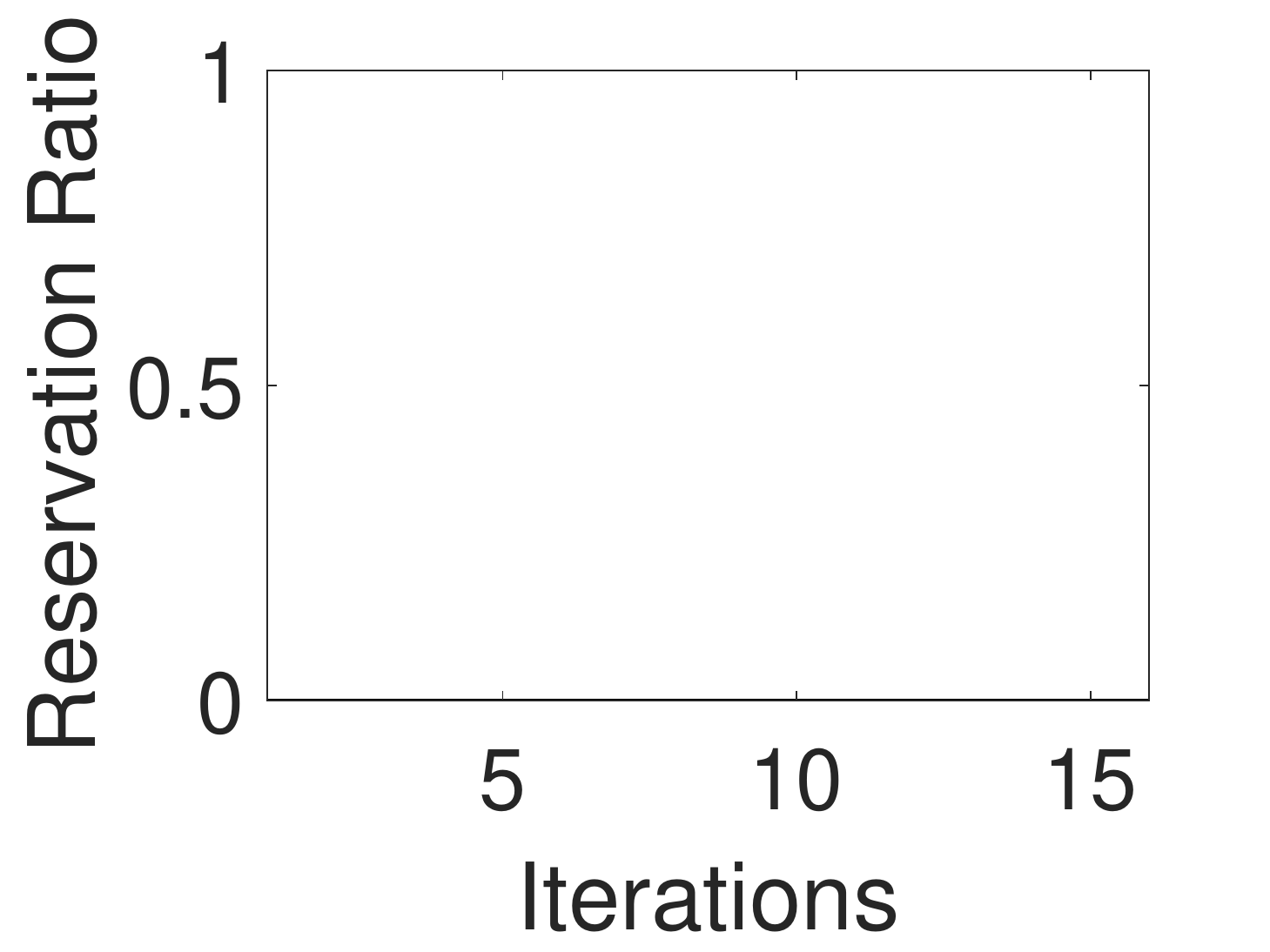}}
		\caption{Reservation ratios of our screening method over the iterations on OCR.}
		\label{fig:scaling-ratios-real}
	\end{center}
\end{figure*}

Figures \ref{fig:reject-ratios-real} and \ref{fig:scaling-ratios-real} report the rejection ratios and reservation ratios of our screening method, respectively. We can see that our approach is very powerful in identifying the irrelevant features. We can identify most all of the irrelevant features and reduce the problem size to nearly zero very quickly. 

The time cost of the training algorithm with and without screening is reported in Table \ref{table:run-time-sync}. We can see that our screening method leads to significant speedups, up to 5.4 times, which is consistent with the rejection ratios and reservation ratios reported above.

\section{Conclusion}
In this paper, we proposed a safe dynamic screening method for sparse CRF to accelerate its training process. Our major contribution is a novel dual optimum estimation developed by carefully studying the strong convexity and the structure of the dual problem. We made the first attempt to  extend the technique used in static screening methods to dynamic screening. In this way, we can absorb the advantages of both static and dynamic screening methods and avoid their shortcomings at the same time. We also developed a detailed screening based on the estimation by solving a convex optimization problem, which can be integrated with the general training algorithms flexibly. An appealing feature of our screening method is that it achieves the speedup without sacrificing any accuracy on the final learned model. To the best of our knowledge, our approach is the first specific screening rule for sparse CRFs. Experiments on  both synthetic and real datasets demonstrate that our approach can achieve significant speedups in training sparse CRF models. 

\bibliographystyle{plain}	
\bibliography{egbib}

\clearpage
\newpage

\appendix
\section{Appendix}
In this supplement, we present the detailed proofs of all the theorems in the main text.
\subsection{Proof of Theorem \ref{theorem:primal-dual-kkt}}
\proof of Theorem \ref{theorem:primal-dual-kkt}:\\
(i): Let $\z_i = -A_i^T \w$, then the primal problem \ref{eqn:primal-problem} is equivalent to 
\begin{align}
&\min_{\w \in \R^d, \z_i \in \R^{|\calY_i|}} P(\w) = \frac{1}{n}\sum_{i=1}^{n}\phi_i(\z_i)+\frac{\alpha\beta}{2}\|\w\|_2^2 + \beta \|\w\|_1, \nonumber\\
&\mbox{s.t. } \z_i = -A_i^T\w, i =1,...,n.\nonumber 
\end{align}
The Lagrangian then becomes 
\begin{align}
&L(\w,\z_1,...,\z_n,\theta_1,...,\theta_n)\nonumber \\
=&\frac{\alpha\beta}{2}\|\w\|_2^2 + \beta \|\w\|_1 + \frac{1}{n}\sum_{i=1}^{n}\phi_i(\z_i) - \frac{1}{n}\sum_{i=1}^{n}\langle \theta_i, \z_i+A_i^T\w \rangle \nonumber \\ 
=&\underbrace{\frac{\alpha\beta}{2}\|\w\|_2^2 + \beta \|\w\|_1 -\langle \frac{1}{n}\sum_{i=1}^{n} A_i \theta_i,\w \rangle }_{:=f_1(\w)} + \underbrace{\frac{1}{n}\sum_{i=1}^{n}\phi_i(\z_i) - \langle \theta_i, \z_i \rangle}_{:=f_2(\z_1,...,\z_n)}\nonumber
\end{align}
We first consider the subproblem $\min_{\w}L(\w,\z_1,...,\z_n,\theta_1,...,\theta_n)$:
\begin{align}
&0 \in \partial_{\w} L(\w,\z_1,...,\z_n,\theta_1,...,\theta_n)= \partial_{\w} f_1(\w) = \alpha\beta \w -  \frac{1}{n}\sum_{i=1}^{n} A_i \theta_i + \beta \partial \|\w\|_1 \Leftrightarrow \nonumber\\
&\frac{1}{n}\sum_{i=1}^{n} A_i \theta_i \in \alpha\beta \w + \beta \partial \|\w\|_1 \Rightarrow \w = 
\frac{1}{\alpha\beta}\calS_{\beta}\big(\frac{1}{n}\sum_{i=1}^{n} A_i \theta_i\big) \label{eqn:sol-w}
\end{align}
By substituting (\ref{eqn:sol-w}) into $f_1(\w)$, we get
\begin{align}
f_1(\w) =&\frac{\alpha\beta}{2}\|\w\|_2^2 + \beta \|\w\|_1 -\langle \alpha\beta \w + \beta \partial \|\w\|_1, \w \rangle \nonumber \\
=& -\frac{\alpha\beta}{2} \|\w\|_2^2\nonumber \\
=& -\frac{1}{2\alpha\beta}\|\calS_{\beta}\big(\frac{1}{n}\sum_{i=1}^{n} A_i \theta_i\big)\|_2^2. \label{eqn:f1-value} 
\end{align}
Then, we consider the problem $\min_{\z_i}L(\w,\z_1,...,\z_n,\theta_1,...,\theta_n)$:
\begin{align}
&0 = \nabla_{\z_i} L(\w,\z_1,...,\z_n,\theta_1,...,\theta_n) = \frac{1}{n}(\nabla \phi_i(\z_i) - \theta_i)\nonumber \\
&\Leftrightarrow [\theta_i]_j = \frac{\exp([\z_i]_j)}{\sum_{k=1}^{|\calY_i|}\exp([\z_i]_k)}, j = 1,...,|\calY_i|. \label{eqn:sol-z}
\end{align}
By substituting (\ref{eqn:sol-z}) into $f_2(\z_1,...,\z_n)$, we get
\begin{align}
&f_2(\z_1,...,\z_n) \nonumber \\
=& \frac{1}{n}\sum_{i=1}^{n} \Big[ \log\Big(\sum_{k=1}^{|\calY_i|}\big([\theta_i]_k \sum_{l=1}^{|\calY_i|}\exp([\z_i]_l)\big)\Big) -\sum_{j=1}^{|\calY_i|}[\theta_i]_j \big(\log([\theta_i]_j) + \log(\sum_{l=1}^{|\calY_i|}\exp([\z_i]_l))\big) \Big]\label{eqn:f2-value-1} \\
=& \frac{1}{n}\sum_{i=1}^{n} \Big[\log \big( \sum_{j=1}^{|\calY_i|}\exp([\z_i]_j) \big)- \sum_{j=1}^{|\calY_i|}[\theta_i]_j \big(\log([\theta_i]_j) -  \log\big(\sum_{l=1}^{|\calY_i|}\exp([\z_i]_l)\big)\Big ] \label{eqn:f2-value-2} \\
=& -\frac{1}{n}\sum_{i=1}^{n}\sum_{j=1}^{|\calY_i|}[\theta_i]_j \big(\log([\theta_i]_j) \nonumber \\
=& \frac{1}{n} \sum_{i=1}^{n} \tilde{H}(\theta_i) \label{eqn:f2-value}
\end{align}
where  $\tilde{H}(\theta_i)=-\sum_{j=1}^{|\calY_i|}[\theta_i]_j \big(\log([\theta_i]_j)$.

The quality (\ref{eqn:f2-value-1}) above comes from the facts that
\begin{align}
\exp([\z_i]_j) = [\theta_i]_j \sum_{k=1}^{|\calY_i|}\exp([\z_i]_k) \mbox{ and } [\z_i]_j = \log([\theta_i]_j) + \log\big(\sum_{k=1}^{|\calY_i|}\exp([\z_i]_k) \big).\nonumber
\end{align}
The equality (\ref{eqn:f2-value-2}) comes from that 
\begin{align}
\sum_{k=1}^{|\calY_i|}[\theta_i]_k = 1, i = 1,...,n. \nonumber 
\end{align}
Combining (\ref{eqn:f1-value}) and (\ref{eqn:f2-value}), we obtain the dual problem:
\begin{align}
&\min_{\theta} D(\theta) = \frac{1}{2\alpha\beta}\|\calS_{\beta}\big(\frac{1}{n}\sum_{i=1}^{n} A_i \theta_i\big)\|_2^2-\frac{1}{n}\sum_{i=1}^{n} \tilde{H}(\theta_i)\nonumber \\
&\mbox{s.t. } \theta = (\theta_1,...,\theta_n) \in \tilde{\Delta}_{|\calY_1|} \times...\times \tilde{\Delta}_{|\calY_n|}, \label{eqn:proof-dual-problem}
\end{align}
where $\tilde{\Delta}_k$ is the $k$-dimensional probability simplex in $\R^k$. 

(ii): From (\ref{eqn:sol-w}) and (\ref{eqn:sol-z}), we  get the KKT conditions:
\begin{align}
&\w^*(\alpha,\beta) = \frac{1}{\alpha\beta}\calS_{\beta}\big(\frac{1}{n}\sum_{i=1}^{n} A_i \theta_i^*(\alpha,\beta)\big)\label{eqn:proof-KKT-1} \\
&\theta^*_i(\alpha,\beta) = \nabla \phi_i(-A_i^T \w^*(\alpha, \beta))\label{eqn:proof-KKT-2}
\end{align}

From \ref{eqn:sol-z}, we have $\sum_{k=1}^{|\calY_i|}[\theta_i]_k = 1, i = 1,...,n.$ Thus, we can present $[\theta_i]_{|\calY_i|}$ by $1- \sum_{j=1}^{|\calY_i|-1}[\theta_i]_j$. Then, we can rewrite the problem \ref{eqn:proof-dual-problem} as:

\begin{align}
&\min_{\theta} D(\theta;\alpha, \beta) = \frac{1}{2\alpha\beta}\|\calS_{\beta}\big(\frac{1}{n}\sum_{i=1}^{n} A_i\circ \theta_i\big)\|_2^2-\frac{1}{n}\sum_{i=1}^{n}H(\theta_i), \nonumber\\
&\mbox{s.t. } \theta = (\theta_1,...,\theta_n) \in \Delta_{|\calY_1|-1} \times...\times \Delta_{|\calY_n|-1}, \nonumber 
\end{align}
where $H(\theta_i) = -\sum_{j=1}^{|\calY_i|-1}[\theta_i]_j\log([\theta_i]_j)-(1-\langle \theta_i, \textbf{1}\rangle)\log(1-\langle \theta_i, \textbf{1}\rangle)$ and $\textbf{1}$ is a vector with all components equal to 1.

And the KKT conditions (\ref{eqn:proof-KKT-1}) and (\ref{eqn:proof-KKT-2}) can be rewritten as: 
\begin{align}
&\w^*(\alpha, \beta) = \frac{1}{\alpha\beta}\calS_{\beta}\big(\frac{1}{n}\sum_{i=1}^{n} A_i\circ \theta_i^*(\alpha, \beta)\big)\nonumber\\
&\tilde{\theta}_i^*(\alpha, \beta) = \nabla \phi_i(-A_i^T\w^*(\alpha, \beta)), \mbox{ with } \tilde{\theta}_i^*(\alpha, \beta) = (\theta_i^*(\alpha, \beta), 1-\langle \theta_i^*(\alpha, \beta), \textbf{1} \rangle),  i = 1,...,n. \nonumber
\end{align} 
The proof is complete. 
\endproof

\subsection{Proof of Lemma \ref{lemma:scaled-problem}}
\proof of Lemma \ref{lemma:scaled-problem}: \\
\textup{(i):} Since $\hcalF \subseteq \calF$, we have $[\w^*(\alpha, \beta)]_{\hcalF} = 0$. Then, the rest components of $\w^*(\alpha, \beta)$ can be recovered by fixing  $[\w]_{\hcalF} = 0$ in problem (\ref{eqn:dual-problem}) and solving the scaled problem:
\begin{align}
\min_{\hat{\w} \in \R^{|\hcalF^c|}} \hat{P}(\hat{\w};\alpha, \beta) = \frac{1}{n}\sum_{i=1}^{n}\phi_i(-{}_{\hcalF^c}[A_i]^T \hat{\w})+ \beta r(\hat{\w}). \nonumber
\end{align}

\textup{(ii): } From the proof of Theorem \ref{theorem:primal-dual-kkt}, we can easily get the dual problem of problem (\ref{eqn:dual-problem-scaled}):
\begin{align}
&\min_{\hat{\theta}} \hat{D}(\hat{\theta};\alpha, \beta) = \frac{1}{2\alpha\beta}\|\calS_{\beta}\big(\frac{1}{n}\sum_{i=1}^{n} {}_{\hcalF^c}[A_i]\circ \hat{\theta}_i\big)\|_2^2-\frac{1}{n}\sum_{i=1}^{n}H(\hat{\theta}_i),\nonumber \\
&\mbox{s.t. } \hat{\theta} = (\hat{\theta}_1,...,\hat{\theta}_n) \in \Delta_{|\calY_1|-1} \times...\times \Delta_{|\calY_n|-1}. \nonumber 
\end{align} 

\textup{(iii): } From the proof of Theorem \ref{theorem:primal-dual-kkt}, we can get the KKT conditions between (\ref{eqn:primal-problem-scaled}) and (\ref{eqn:dual-problem-scaled}):

\begin{align}
&\hat{\w}^*(\alpha, \beta) = \frac{1}{\alpha\beta}\calS_{\beta}\big(\frac{1}{n}\sum_{i=1}^{n} {}_{\hcalF^c}[A]_i\circ \hat{\theta}_i^*(\alpha, \beta)\big), \nonumber \\
&(\hat{\theta}_i^*(\alpha, \beta), 1-\langle \hat{\theta}_i^*(\alpha, \beta), \textbf{1} \rangle) = \nabla \phi_i(-{}_{\hcalF^c}[A]_i^T\hat{\w}^*(\alpha, \beta)), i = 1,...,n. \nonumber 
\end{align} 

Then, to proof $\theta^*(\alpha,\beta) = \hat{\theta}^*(\alpha,\beta)$, we just need to verify that $\hat{\w}^*(\alpha,\beta)$ and $\theta^*(\alpha, \beta)$ satisfy the KKT conditions above. 

Actually, from \ref{eqn:KKT-1} and $[\w^*(\alpha,\beta)]_{\hcalF^c} = \hat{\w}^*(\alpha, \beta)$, we have
\begin{align}
\hat{\w}^*(\alpha, \beta) = \frac{1}{\alpha\beta}\calS_{\beta}\big(\frac{1}{n}\sum_{i=1}^{n} {}_{\hcalF^c}[A]_i\circ \theta_i^*(\alpha, \beta)\big).\nonumber
\end{align}

From \ref{eqn:KKT-2} and the fact that $[\w^*(\alpha,\beta)]_{\hcalF^c} = \hat{\w}^*(\alpha, \beta)$ and $[\w^*(\alpha,\beta)]_{\hcalF}=\mathbf{0}$, we have,
\begin{align}
(\theta_i^*(\alpha, \beta), 1-\langle \theta_i^*(\alpha, \beta), \textbf{1} \rangle) = \nabla \phi_i(-{}_{\hcalF^c}[A]_i^T\hat{\w}^*(\alpha, \beta)), i = 1,...,n. \nonumber 
\end{align} 

Therefore, $\theta^*(\alpha,\beta) = \hat{\theta}^*(\alpha,\beta)$. 

The proof is complete. 
\endproof

\subsection{Proof of Theorem \ref{theorem:beta-max}}
\proof of Theorem \ref{theorem:beta-max}: \\
We just need to verify that, when $\alpha>0, \beta>\beta_{\max}$, 	$\w^*(\alpha, \beta) = \mathbf{0} \mbox{ and } \theta^*(\alpha,\beta) = \frac{1}{|\calY_i|} \mathbf{1}, i = 1,...,n$ satisfy  the conditions (\ref{eqn:KKT-1}) and (\ref{eqn:KKT-2}). 

To be precise, since $\beta> \beta_{\max} = \frac{1}{n} \|\sum_{i=1}^{n} A_i\frac{ \mathbf{1}}{|\calY_i|}\|_{\infty}$, we have\\
\begin{align}
\frac{1}{n} \|\sum_{i=1}^{n} A_i\circ \theta^*(\alpha,\beta)\|_{\infty} \leq \beta,
\end{align}
Therefore,
\begin{align}
\frac{1}{\alpha\beta}\calS_{\beta}\big(\frac{1}{n}\sum_{i=1}^{n} A_i\circ \theta_i^*(\alpha, \beta)\big) = \mathbf{0} = \w^*(\alpha,\beta). \nonumber 
\end{align}
Thus, the condition \ref{eqn:KKT-1} holds.

Since $\nabla \phi(\mathbf{0}) = \frac{1}{|\calY_i|}$, it is straightforward to verify that the condition \ref{eqn:KKT-2} holds. 

The proof is complete. 
\endproof

\subsection{Proof of Lemma \ref{lemma:strong-convex}}
\proof of Lemma \ref{lemma:strong-convex}:\\
This is equivalent to prove that $\hat{D}(\hat{\theta};\alpha,\beta)$ is $\frac{1}{n}$-strongly convex. Since $\frac{1}{2\alpha\beta}\|\calS_{\beta}\big(\frac{1}{n}\sum_{i=1}^{n} A_i \circ \hat{\theta}_i\big)\|_2^2$ is convex, we just need to verify that $h(\hat{\theta}) =-\frac{1}{n}\sum_{i=1}^{n}H(\hat{\theta}_i)$ is $\frac{1}{n}$-strongly convex.

Firstly, we need to give the gradient of $h(\hat{\theta})$ as follows: 
\begin{align}
&[\nabla h(\hat{\theta})] = -\frac{1}{n}[\nabla H(\hat{\theta}_1);...; \nabla H(\hat{\theta}_n)], \nonumber \\
&\mbox{where } [\nabla H(\hat{\theta}_i)]_j =-\log([\hat{\theta}_i]_j) + \log(1-\sum_{j=1}^{|\calY_i|-1}[\hat{\theta}_i]_j), i=1,...,n \mbox{ and } j = 1,...,|\calY_i|-1.\nonumber
\end{align}
Thus, we have
\begin{align}
\nabla^2 h(\hat{\theta}) = \begin{pmatrix}
-\frac{1}{n}\nabla^2H(\hat{\theta}_1)&&&0\\ &-\frac{1}{n}\nabla^2H(\hat{\theta}_2)& &\\ &&...& \\ 0&& &-\frac{1}{n}\nabla^2H(\hat{\theta}_n) \\
\end{pmatrix} \nonumber 
\end{align}
$[\nabla^2 H(\hat{\theta}_i)]_{jk} =\begin{cases}
-\frac{1}{[\hat{\theta}_i]_j}- \frac{1}{1-\sum_{j=1}^{|\calY_i|-1}[\hat{\theta}_i]_j} \mbox{ if } j=k,\\
- \frac{1}{1-\sum_{j=1}^{|\calY_i|-1}[\hat{\theta}_i]_j}  \mbox{ otherwise.}
\end{cases}$ with $i = 1,...,n$, $j=1,...,|\calY_i|-1$ and $k=1,...,|\calY_i|-1$.

Below, we will prove that $-\nabla^2 H(\hat{\theta}) > I$ where $I$ is a identity matrix in $\R^{(|\calY_i|-1)\times (|\calY_i|-1)}$.  

We notice that $-\nabla^2 H(\hat{\theta})$ can be rewritten as a sum of two matrix:
\begin{align}
-\nabla^2 H(\hat{\theta}) = \frac{1}{1-\sum_{j=1}^{|\calY_i|-1}[\hat{\theta}_i]_j}\mathbf{ones}(|\calY_i|-1,|\calY_i|-1) + \textup{diag}(\frac{1}{[\hat{\theta}_i]_1},...,\frac{1}{[\hat{\theta}_i]_{|\calY_i|-1}}), \nonumber
\end{align}
where $\mathbf{ones}(|\calY_i|-1,|\calY_i|-1)$ is a matrix in $\R^{(|\calY_i|-1)\times (|\calY_i|-1)}$ with all elements are 1. Hence, from Weyl theorem, we have
\begin{align}
\textup{eig}_{\min}\Big(-\nabla^2 H(\hat{\theta})\Big) &\geq \textup{eig}_{\min}\Big(\frac{1}{1-\sum_{j=1}^{|\calY_i|-1}[\hat{\theta}_i]_j}\mathbf{ones}(|\calY_i|-1,|\calY_i|-1) \Big)\nonumber \\
& + \textup{eig}_{\min}\Big( \textup{diag}(\frac{1}{[\hat{\theta}_i]_1},...,\frac{1}{[\hat{\theta}_i]_{|\calY_i|-1}}) \Big)\nonumber \\
&= 0 + \min_j \frac{1}{[\hat{\theta}_i]_j} > 1. \nonumber 
\end{align}
where $\textup{eig}_{\min}(\cdot)$ presents the minimal eigenvalue of a matrix. Hence, $-\nabla^2 H(\hat{\theta}) > I$. 

Therefore, $\hat{D}(\hat{\theta};\alpha,\beta)$ is $\frac{1}{n}$-strongly convex. 

The proof is complete. 
\endproof

\subsection{Proof of Lemma \ref{lemma:dual-estimation}}
\proof of Theorem \ref{lemma:dual-estimation}:\\
\textup{(i): } We define $f(\theta) = \hat{D}(\theta;\alpha,\beta)- \langle\nabla \hat{D}(\theta_0;\alpha, \beta), \theta  \rangle $ Since $\hat{D}(\hat{\theta};\alpha,\beta)$ is $\frac{1}{n}$-strongly convex, we can get $f(\theta)$ is $\frac{1}{n}$ strongly convex.

Since $\nabla f(\theta_0) = 0$, then for any $\theta$, we have
\begin{align}
f(\theta_0) =& \min_{\x} f(\x) \geq \min_{\x} f(\theta) + \langle \nabla f(\theta), \x-\theta \rangle + \frac{1}{2n} \|\x-\theta\|_2^2\nonumber\\ 
=&f(\theta) - \frac{n}{2}\|\nabla f(\theta)\|^2\nonumber
\end{align}
Let $\theta = \hat{\theta}^*(\alpha, \beta)$ and $\theta_0 = \hat{\theta}$, the inequality above implies that $I_1(\hat{\theta})\geq 0$. 

$I_2(\hat{\theta})\geq 0$ comes from the fact that $\hat{\theta}^*(\alpha, \beta)$ is the optimal solution of $\hat{D}$. 

\textup{(ii)}: $I_1(\hat{\theta}) \geq 0$ and $ I_2(\hat{\theta})\geq 0$ comes from the continuity of $\hat{D}(\hat{\theta};\alpha,\beta)$ and $\nabla \hat{D}(\hat{\theta};\alpha,\beta)$. 

From Lemma \ref{lemma:strong-convex}, we have
\begin{align}
&\|\hat{\theta}^*(\alpha, \beta)- \hat{\theta}_1\|^2 \nonumber \\
\leq & 2n\Big(D(\hat{\theta}^*(\alpha, \beta);\alpha,\beta)-D(\hat{\theta}_1;\alpha,\beta) - \langle \nabla D(\hat{\theta}_1;\alpha,\beta), \theta^*(\alpha, \beta)-\hat{\theta}_1\rangle \Big)\nonumber\\
\leq & 2n \Big(D(\hat{\theta}_2;\alpha,\beta)-D(\hat{\theta}_1;\alpha,\beta) - \langle \nabla D(\hat{\theta}_1;\alpha,\beta), \theta^*(\alpha, \beta)-\hat{\theta}_1\rangle \Big) \nonumber 
\end{align}
Thus, we can get 
\begin{align}
&\|\hat{\theta}^*(\alpha, \beta)- \hat{\theta}_1\|^2 + 2n\langle \nabla D(\hat{\theta}_1;\alpha,\beta), \theta^*(\alpha, \beta)\rangle \nonumber\\
\leq& 2n \Big(D(\hat{\theta}_2;\alpha,\beta)-D(\hat{\theta}_1;\alpha,\beta) +\langle \nabla D(\hat{\theta}_1;\alpha,\beta), \hat{\theta}_1\rangle \Big)\nonumber
\end{align}
Therefore, 
\begin{align}
&\|\hat{\theta}^*(\alpha, \beta)- \big(\hat{\theta}_1-n\nabla D(\hat{\theta}_1;\alpha,\beta)
\big)\|^2\nonumber \\
\leq & 2n \Big(D(\hat{\theta}_2;\alpha,\beta)-D(\hat{\theta}_1;\alpha,\beta) +\langle \nabla D(\hat{\theta}_1;\alpha,\beta), \hat{\theta}_1\rangle \Big)\nonumber \\
&-2n\langle \nabla D(\hat{\theta}_1;\alpha,\beta), \hat{\theta}_1\rangle + n^2 \|\nabla D(\hat{\theta}_1;\alpha,\beta)\|_2^2\nonumber \\
=& 2n \Big(D(\hat{\theta}_2;\alpha,\beta)-D(\hat{\theta}_1;\alpha,\beta) \Big) + n^2 \|\nabla D(\hat{\theta}_1;\alpha,\beta)\|_2^2\nonumber\\
=&2n \Big(D(\hat{\theta}_2;\alpha,\beta)-D(\hat{\theta}^*(\alpha, \beta);\alpha,\beta) \Big) +\Big( n^2 \|\nabla D(\hat{\theta}_1;\alpha,\beta)\|_2^2  -D(\hat{\theta}_1;\alpha,\beta)+ D(\hat{\theta}^*(\alpha, \beta);\alpha,\beta)\Big)\nonumber\\
=&2nI_2(\hat{\theta}_2) + nI_1(\hat{\theta}_1)\nonumber
\end{align}

The proof is complete. 
\endproof
\subsection{Proof of Theorem \ref{theorem:dual-estimation}}
\proof of Theorem \ref{theorem:dual-estimation}:\\
From Theorem \ref{theorem:primal-dual-kkt}, we have $\hat{\theta}_i^*(\alpha, \beta) \in \Delta_{|\calY_i|-1}$, which implies that 
\begin{align}
\langle \hat{\theta}_i^*(\alpha, \beta), \mathbf{1} \rangle \leq 1, i =1,...,n.\nonumber 
\end{align}
Therefore, $\hat{\theta}_i \in \calH_i, i = 1,...,n$. 

From Theorem \ref{lemma:dual-estimation}, we have $\hat{\theta}^*(\alpha, \beta)\in \calB$. Thus, we finally have
\begin{align}
\hat{\theta}^*(\alpha, \beta)\in \Theta = \calB\cap \calH_1\cap \calH_2\cap...\cap \calH_n.\nonumber
\end{align}
The proof is complete. 

\endproof
\subsection{Proof of Theorem \ref{theorem:closed-form-solution}}
\proof of Theorem \ref{theorem:closed-form-solution}:\\
Let denote the optimal value as $T$, that is,
\begin{align}
&T = \min \langle \a, \b \rangle \nonumber\\
&s.t. \|\a-\a_0\|_2^2 \leq r^2 \mbox{ and } \langle \a, \p \rangle \leq c. \nonumber
\end{align}

If $\b=\mathbf{0}$, then $T=0$.

If $\b \neq \mathbf{0}$, then we have the following Lagrangian with dual variables $\lambda_1\in \R$ and $\lambda_2\geq 0$:
\begin{align}
&L(\a, \lambda_1, \lambda_2) = \langle \a, \b\rangle + \lambda_1(\| \a- \a_0\|^2-r^2) + \lambda_2 (\langle\p, \a \rangle-c),\nonumber \\
&L(\lambda_1, \lambda_2) = \min_{\a} \langle \a, \b\rangle + \lambda_1(\| \a- \a_0\|^2-r^2) + \lambda_2 (\langle\p, \a \rangle-c),\nonumber\\
& \nabla_{\a}L(\a, \lambda_1,\lambda_2) = \b + 2 \lambda_1\a - 2\lambda_1 \a_0+ \lambda_2 \p = 0\nonumber
\end{align}
(1) If $\lambda_1\neq 0$, we have 
\begin{align}
\a = \frac{-\b+2\lambda_1 \a_0 -\lambda_2 \p}{2\lambda_1}\nonumber
\end{align}
Plug $\a$ above into $L(\a, \lambda_1, \lambda_2)$, we have
\begin{align}
L(\lambda_1,\lambda_2) = - \frac{1}{4\lambda_1}\|\b + \lambda_2 \p\|_2^2 + \langle \b +\lambda_2 \p, \a_0 \rangle - (\lambda_1r^2 +\lambda_2c)\nonumber 
\end{align}

(2) If $\lambda_1 = 0$, we have 
\begin{align}
L(\lambda_1,\lambda_2) = \min_{\a} \big(\langle \b+ \lambda_2\p,\a \rangle-\lambda_2c  \big)\nonumber
\end{align}
i): $\lambda_2 \neq \frac{\|\b\|_2}{\|\p\|_2}$, we have $L(\lambda_1,\lambda_2) = -\infty$. 

ii): $\lambda_2 \neq \frac{\|\b\|_2}{\|\p\|_2}$, then $L(\lambda_1,\lambda_2) =-\frac{\|\b\|_2}{\|\p\|_2}c $.

Therefore, the dual problem can be written as follows:
\begin{align}
L(\lambda_1,\lambda_2) = \begin{cases}
- \frac{1}{4\lambda_1}\|\b + \lambda_2 \p\|_2^2 + \langle \b +\lambda_2 \p, \a_0 \rangle - (\lambda_1r^2 +\lambda_2c), \mbox{ if } \lambda_1 \neq  0, \lambda_2\geq 0,\\
-\frac{\|\b\|_2}{\|\p\|_2}c, \mbox{ if } \lambda_1 = 0, \lambda_2 = \frac{\|\b\|_2}{\|\p\|_2}\\
-\infty, \mbox{ if } \lambda_1 =0,  \lambda_2 \neq \frac{\|\b\|_2}{\|\p\|_2}, \lambda_2 \geq 0.
\end{cases}\nonumber
\end{align}
Our problem becomes to solve $\max_{\lambda_1, \lambda_2\geq 0}L(\lambda_1, \lambda_2)$. When $\langle \p, \b \rangle > -\|\p\|_2 \|\b\|_2$, only $\lambda_1>0$ and $\lambda_2\geq 0$ will make sense. So. we solve the problem below via KKT conditions:
\begin{align}
\min_{\lambda_1>0, \lambda_2\geq 0} -L(\lambda_1, \lambda_2) = \frac{1}{4\lambda_1}\|\b + \lambda_2 \p\|_2^2 -\langle \b +\lambda_2 \p, \a_0 \rangle + (\lambda_1r^2 +\lambda_2c). \nonumber
\end{align}
The KKT conditions are:
\begin{align}
&-\frac{1}{4\lambda_1^2}\|\b + \lambda_2 \p\|_2^2 + r^2 - \mu_1 = 0 \label{eqn:proof-kkt1}\\
&\frac{1}{2\lambda_1}\langle\p, \b + \lambda_2\p \rangle-\langle \p, \a_0 \rangle+ c- \mu_2 = 0\label{eqn:proof-kkt2}\\
&\mu_1 \lambda_1 = 0\label{eqn:proof-kkt3} \\
&\mu_2\lambda_2 = 0\label{eqn:proof-kkt4}
\end{align}
Since $\lambda_1>0$, we have $\mu_1=0$, so $\lambda_1 = \frac{\|\b + \lambda_2\p\|_2}{2r}$. Plug it into (\ref{eqn:proof-kkt1}), we have 
\begin{align}
\frac{\langle\p, \b + \lambda_2\p \rangle}{\|\p\|_2 \|\b + \lambda_2\p\|_2 } = \frac{\langle \p, \a_0 \rangle-c}{r\|\p\|_2} + \frac{\mu_2}{r\|\p\|_2}. \nonumber 
\end{align}
When $\pm \langle\p, \b \rangle \neq \|\p\|_2 \|\b\|_2$:\\
The right part will increase to 1 when $\lambda_2$ increases. 

1) If $\frac{\langle\p, \b \rangle}{\|\p\|_2 \|\b\|_2} >  \frac{\langle \p, \a_0 \rangle-c}{r\|\p\|_2} $, then 
\begin{align}
\frac{\langle \p, \a_0 \rangle-c}{r\|\p\|_2} + \frac{\mu_2}{r\|\p\|_2} = \frac{\langle\p, \b + \lambda_2\p \rangle}{\|\p\|_2 \|\b + \lambda_2\p\|_2 }> \frac{\langle \p, \a_0 \rangle-c}{r\|\p\|_2}. \nonumber 
\end{align}
Since $\lambda \geq 0$, which implies $\mu_2>0$. Thus $\lambda_2 = 0$. 

Therefore, we have $\lambda_1 = \frac{\|\b\|_2}{2r}, \lambda_2 = 0$. 

2) If $\frac{\langle\p, \b \rangle}{\|\p\|_2 \|\b\|_2} =  \frac{\langle \p, \a_0 \rangle-c}{r\|\p\|_2} $ and $\lambda_2 >0$, then $\mu_2 =0$. Therefore, 
\begin{align}
\frac{\langle \p, \a_0 \rangle-c}{r\|\p\|_2}=\frac{\langle\p, \b \rangle}{\|\p\|_2 \|\b\|_2} \Rightarrow \lambda_2 = 0, \mbox{ contradict}.  \nonumber
\end{align}
So, $\lambda_2 = 0, \lambda_1 = \frac{\|\b\|_2}{2r}$. 

The above two cases 1) and 2) all leads  to that 
\begin{align}
-L(\lambda_1, \lambda_2) = r\|\b\|- \langle\b,\a_0 \rangle. \nonumber 
\end{align}

3) If $ \frac{\langle\p, \b \rangle}{\|\p\|_2 \|\b\|_2} <  \frac{\langle \p, \a_0 \rangle-c}{r\|\p\|_2} $:

If $\lambda_2=0$, $ \frac{\langle\p, \b \rangle}{\|\p\|_2 \|\b\|_2} =  \frac{\langle \p, \a_0 \rangle-c}{r\|\p\|_2} +  \frac{\mu_2}{r\|\p\|_2}.$ $\Rightarrow \mu_2 <0$, contradiction of $\mu_2 \geq 0$. 

Therefore $\lambda_2>0$ and $\mu_2 = 0$. 

Thus, we have $\lambda_1 = \frac{\|\b + \lambda_2\p\|_2}{2r}$, $ \frac{\langle \p, \a_0 \rangle-c}{r\|\p\|_2} = \frac{\langle\p, \b + \lambda_2\p \rangle}{\|\p\|_2 \|\b + \lambda_2\p\|_2 }> \frac{\langle\p, \b \rangle}{\|\p\|_2 \|\b\|_2}$ and $\frac{\langle \p, \a_0 \rangle-c}{r\|\p\|_2}<1$. 

Let $\frac{\langle \p, \a_0 \rangle-c}{r\|\p\|_2}=d$, we have 
\begin{align}
&\lambda_2^2 \|\p\|^4 d^2 + 2d^2 \|\p\|_2^2 \langle\b, \p \rangle \lambda_2 +d^2 \|\p\|_2^2 \|\b\|_2^2\nonumber\\
=&\lambda_2^2\|\p\|_2^4 + 2 \langle\p, \b \rangle\|\p\|_2^2 \lambda_2 + (\langle\p, \b \rangle)^2 \nonumber
\end{align}
Therefore, 
\begin{align}
\lambda_2 = \frac{-2(1-d^2)\|\p\|_2^2\langle\p, \b \rangle\|\p+ \sqrt{\Delta} }{2\|\p\|_2^4(1-d^2)}. 
\end{align}

4) When $\|\p\|_2\|\b\|_2 = \langle\p, \b \rangle$:\\
We have $\lambda_1 = \frac{\|\b + \lambda_2\p\|_2}{2r}$, $ \frac{\langle\p, \b + \lambda_2\p \rangle}{\|\p\|_2 \|\b + \lambda_2\p\|_2 } = \frac{\langle \p, \a_0 \rangle-c}{r\|\p\|_2}+ \frac{\mu_2}{r\|\p\|_2} = d + \frac{\mu_2}{r\|\p\|_2}$ and $\frac{\langle\p, \b + \lambda_2\p \rangle}{\|\p\|_2 \|\b + \lambda_2\p\|_2 }=1$. Therefore, $ d + \frac{\mu_2}{r\|\p\|_2}=1$. 

If $d<1$, then $\mu_2 = (1-d)r\|\p\|_2 > 0 \Rightarrow \lambda_2 = 0 \Rightarrow \lambda_1 = \frac{\|\b\|_2}{2r}$.

If $d=1$, then $\mu_2=0$. 

Hence, $-L(\lambda_1, \lambda_2)= \|\b\|_2 r - \langle\b, \a_0 \rangle$.

5) Now, we consider the case where $-\|\p\|_2 \|\b\|_2 = \langle\p, \b \rangle$. 
So, $\b = - \bar{k} \p$ with $\bar{k} = \frac{\|\b\|_2}{\|\p\|_2}$. Then, the dual problem becomes
\begin{align}
L(\lambda_1,\lambda_2) = \begin{cases}
- \frac{1}{4\lambda_1}\|\b + \lambda_2 \p\|_2^2 + \langle \b +\lambda_2 \p, \a_0 \rangle - (\lambda_1r^2 +\lambda_2c), \mbox{ if } \lambda_1 \neq  0, \lambda_2\geq 0,\\
-\frac{\|\b\|_2}{\|\p\|_2}c, \mbox{ if } \lambda_1 = 0, \lambda_2 = \frac{\|\b\|_2}{\|\p\|_2}\\
-\infty, \mbox{ if } \lambda_1 =0,  \lambda_2 \neq \frac{\|\b\|_2}{\|\p\|_2}, \lambda_2 \geq 0.
\end{cases}\nonumber
\end{align}
Since $\b = - \bar{k} \p$, then:
\begin{align}
&-L_1(\lambda_1, \lambda_2) = \frac{1}{4\lambda_1}(\lambda_2 - \bar{k})^2\|\p\|_2^2 - (\lambda_2-\bar{k})\langle \p, \a_0 \rangle + (\lambda_1r^2 +\lambda_2c)\nonumber \\
&-L_2(\lambda_1,\lambda_2) = \frac{\|\b\|_2}{\|\p\|_2}c. \nonumber
\end{align}
Therefore $\min(-L(\lambda_1, \lambda_2)) = \min \{-L_1(\lambda_1, \lambda_2),\frac{\|\b\|_2}{\|\p\|_2}c  \}$. 

Since $\lambda_1 = \frac{\|\b + \lambda_2\p\|_2}{2r}$ and $ \frac{\langle\p, \b + \lambda_2\p \rangle}{\|\p\|_2 \|\b + \lambda_2\p\|_2 }  = d + \frac{\mu_2}{r\|\p\|_2}$, we have 
\begin{align}
\frac{\langle\p, (\lambda_2-\bar{k}) \p \rangle}{\|\p\|_2 \|(\lambda_2-\bar{k})\p\|_2 }  = d + \frac{\mu_2}{r\|\p\|_2}\in \{-1,1\} \nonumber
\end{align}

When $d=1$, then $\mu_2 = 0, \lambda_2 > \bar{k}$. In this case $-L(\lambda_1,\lambda_2)= \bar{k}c$. 

When $d = -1$, then $\begin{cases}
\mu_2 =0, \lambda_2 < \bar{k} \Rightarrow -L(\lambda_1,\lambda_2)= \bar{k}c. \\
\mu_2 = 2r\|\p\|_2, \lambda_2 > \bar{k}>0, \mbox{ contradict}
\end{cases} $

When $d\in (-1,1)$, then $\begin{cases}
\lambda_2 < \bar{k}, \mu_2 < 0 \mbox{ if } d + \frac{\mu_2}{r\|\p\|_2} = -1 \mbox{ contradict}\\
\lambda_2 > \bar{k}, \mu_2 > 0 \mbox{ if } d + \frac{\mu_2}{r\|\p\|_2} = 1\mbox{ contradict}
\end{cases}$

The proof is complete.
\endproof
\subsection{Proof of Theorem \ref{theorem:screeing-rule}}
\proof of Theorem \ref{theorem:screeing-rule}:\\
\textup{(i): }It can be obtained from rule \ref{eqn:rule-R}. \\
\textup{(ii): }It is from the definition of $\hcalF$. 

The proof is complete. 

\endproof	
\end{document}